\documentclass[journal]{IEEEtran}
\pdfoutput=1
\usepackage{blindtext}
\usepackage{graphicx}
\usepackage{cite}
\usepackage[cmex10]{amsmath}
\usepackage{url}
\begin{document}
\title{Deep Learning in Customer Churn Prediction: Unsupervised Feature Learning on Abstract Company Independent Feature Vectors}%
\author{Philip~Spanoudes, Thomson~Nguyen \\ Framed~Data Inc, New York University,  and the Data~Science~Institute at Lancaster~University \\ \emph{philip@framed.io}, \emph{thomson.nguyen@cs.nyu.edu}}
\maketitle
\begin{abstract}
As companies increase their efforts in retaining customers, being able to predict accurately ahead of time, whether a customer will churn in the foreseeable future is an extremely powerful tool for any marketing team. The paper describes in depth the application of Deep Learning in the problem of churn prediction. Using abstract feature vectors, that can generated on any subscription based company's user event logs,  the paper proves that through the use of the intrinsic property of Deep Neural Networks (learning secondary features in an unsupervised manner), the complete pipeline can be applied to any subscription based company with extremely good churn predictive performance. Furthermore the research documented in the paper was performed for Framed Data (a company that sells churn prediction as a service for other companies) in conjunction with the Data Science Institute at Lancaster University, UK. This paper is the intellectual property of Framed Data.
\end{abstract}

\section{Introduction}
Various markets across the world are becoming increasingly more saturated, with more and more customers swapping their registered services between competing companies. Thus companies have realized that they should focus their marketing efforts in customer retention rather than customer acquisition. In fact, studies have shown that the funds a company spends in attempting to gain new customers is far greater than the funds it would spend if it were to attempt to retain its customers \cite{hadden}. Customer retention strategies can be targeted on high-risk customers that are intending to discontinue their custom or move their custom to another service competitor. This effect of customer loss is better known as customer churn. Thus accurate and early identification of these customers is critical in minimizing the cost of a company’s overall retention marketing strategy.
\par Through the use of machine learning, Framed are able to identify high-risk customers before they churn. This churn assessment is performed monthly for a specific company so that the company can subsequently apply a targeted marketing strategy in order to retain these customers. This high-risk customer identification methodology is sold as a service to various companies that are interested in forming more advanced customer retention strategies.

\subsection{Current Machine Learning Pipeline at Framed}
At the moment, Framed are using a very modern and advanced classification machine learning algorithm knows as the Random Forest algorithm. This is an ensemble classifier and thus has the distinct advantage of not over-fitting its generated model parameters due the Law of Large Numbers \cite{breiman}. Like most conventional machine learning algorithms, Random Forests’ performance in predicting is highly dependent on the features that it is given. Without the capability of engineering its own features to be able to better capture variance present in the data (which would ultimately increase prediction accuracies), a lot of time is spent by Framed in generating secondary features that can do just that. This derivation and generation of meaningful secondary features becomes a struggle when this needs to happen for each and every company Framed provides its service to. This is because each company has its own unique features that exhibit their own variances and dependencies. Fig 1 shows a very basic overview of the operations that take place within Frameds' machine learning pipeline. \par
\begin{figure}
\begin{center}
	\includegraphics[width=3.5in]{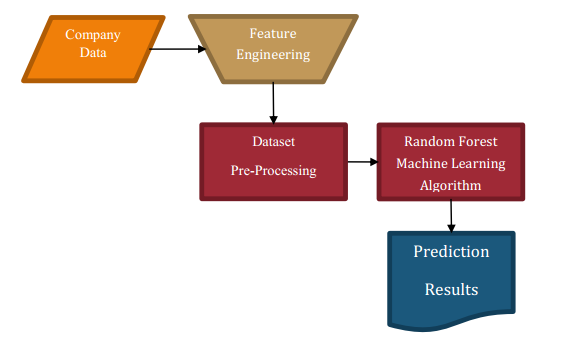}
	\caption{Machine Learning Pipeline at Framed}
\end{center}
\end{figure}
\par Studies have shown that the performance of almost all machine learning algorithms are severely affected by the representation that is used to describe the data they are processing (features) \cite{bengio:dl_representations,bengio:representation_learning}. This is because different features can get entangled with other features and thus would hide some explanatory factors that would describe some of the variation within the data inputs. This phenomenon is also known as the curse of dimensionality \cite{arel}. For that reason most of the effort when designing machine learning systems goes into the feature engineering phase, where machine learning practitioners need to generate new features from the data that would allow the machine learning algorithms to produce adequate results \cite{bengio:dl_representations}. Thus companies like Framed will spend a lot of time and effort in feature engineering in order to optimize their machine learning classifiers for a specific problem. This is especially important when dealing with high-dimensional data. Furthermore, most human generated features usually end up being sub-optimal as most of the time they are either over-specified or incomplete \cite{castanedo}.

\subsection{Dissolving Feature Engineering through Deep Learning}
Advances in the field of Neural Networks and recent increases in computing performance, have allowed for the development of large scale neural networks with more than a single hidden layer. This allowed deep neural networks to propagate the weights of each of their layers to the next. The effects of this ability was that the networks were able to decompose the complexities within the given data by generating abstract data features in an unsupervised manner in each of their hidden layers \cite{bengio:representation_learning,bengio:deep_architectures}. This gave new life to predictive modelling on high-dimensional datasets with very noisy data (image recognition, automatic speech recognition etc.) as the unsupervised abstract features were able to capture the most important variances within the data and thus ignore any variance that did not affect the result variable \cite{le,deng:speech,deng:dlmethods}. \par
This inherent ability have made Deep Neural Networks (DNN’s) excellent tools in pattern recognition. Since churn prediction is the analysis of user behavioural patterns, the application of DNN’s in this domain could definitely be beneficial not only in terms of prediction accuracies but also in eliminating manual feature engineering as a required step.

\subsection{Project Aim and Objectives}
From the presented limitations of human feature engineering in conventional machine learning algorithms which are currently employed at Framed, the project will seek to investigate and apply a deep learning architecture in Frameds’ machine learning pipeline. The deep architecture would allow for unsupervised feature learning which in practice should allow the company to bypass the feature engineering step for any company data they receive. Ideally the deep architecture should also increase the company’s prediction accuracy. \par
Most machine learning implementations are performed and tuned on specific company data based on the company’s business model. Thus input vectors and output labels are pre- processed based on these factors and therefore the representations generated are very task specific. In order to be able to apply a deep learning architecture for any kind of data Framed deals with, the representations need to be abstracted and simplified while also capturing user behavioural patterns. The research conducted will give insight to how well the inherent unsupervised feature engineering ability of DNN’s performs across companies when presented with abstract user behavioural input vectors. \par
The general project aim can be broken down into objectives. The objectives describe the aim of the project in further detail and the collective completion of these objectives should provide reference as to how successfully the projects aim has been satisfied during evaluation.\par
\paragraph{Generate an Encompassing Data Representation Architecture for Deep Learning Prediction}
One of the core issues with applying deep learning architectures in any problem scenario is to generate a specialized data representation architecture. This data representation architecture should structure the data in such a way as to reduce dimensionality while upholding a high-resolution representation of the underlying data features. Due to the fact that features change according to the problem scenario (i.e. features change according to the service a company carries out), an encompassing data representation architecture needs to be developed that can be applied across different companies regardless of the data features each company uses. Creating a generic data representation to encompass different company features is quite novel and its success could be quantified by how well the deep learning architecture performs across companies.
\paragraph{Implement an Appropriate Deep Learning Architecture for Churn Prediction}
The deep learning architecture implemented should be able to employ the unsupervised feature learning ability of deep neural networks. This is critical as this would ensure that the general aim of the project, of avoiding human feature engineering, is satisfied. Furthermore the deep architecture should employ techniques to generalize well across different months, without a lot of variance in prediction accuracy across months. Ideally the deep learning architecture should perform better in terms of prediction accuracy against the currently employed machine learning algorithm at Framed for the specific companies that will be investigated. This is not a requirement as the general aim of this research is to avoid manual feature engineering but it will definitely be a positive result if this is achieved.
\subsection{Paper Overview}
The remainder of the report is structured as follows. In Section II, an overview of churn prediction and its applications are presented as well as an in depth overview of the research that was conducted in understanding deep learning and the Spark computational cluster. The Methodology that was followed to develop the proposed data representation algorithm and the deep learning architecture are presented in Section III. Section IV covers the steps taken in evaluating and analysing the prediction results of the proposed deep learning architecture. The prediction results are discussed in Section V and in Section VI we draw our conclusions.

\section{Background Research and Related Work}
\subsection{Churn Prediction Applications}
Companies are becoming increasingly more aware of the fact that retaining existing customers is the best marketing strategy to follow in order survive in industry \cite{kimparkjeong}. In order to be able to apply these marketing strategies, customers that are likely to move their custom to a competitor need to be identified. The effect of customer abandoning their custom with a service provider is better known as churn. Applying retention strategies becomes even more important in the case of mature businesses whose customer base has reached its peak and thus retaining customers is of upmost importance.\par
The reasons behind why customers might want to discontinue their custom with a company can vary. This can be divided into two types of churn: incidental and deliberate churn \cite{kimyoon}. Sometimes customers are forced into dropping their service with a company due to life circumstances. This is known as incidental churn. Some examples include customer relocation to areas where the company does not provide service to, or even changes in a customer’s financial status such that he/she can no longer afford to stay with a company. Deliberate churn describes the effect of a customer churning due to the customer deciding to move their custom to a competitor. Reasons behind this can range from a competitor offering a latest product, a competitor having better prices for the same service or even the customer’s bad experience with technical support (call centres).\par
From the reasons presented above it becomes clear that it is of great importance for a company to understand its customers in order for it to evolve its business strategy. Thus identifying customers who are about to churn becomes not just important in terms of retaining customers but also in terms of gathering business intelligence. As a response in tackling this problem companies have turned to predictive modelling techniques to assist in the identification of these customers. Numerous different machine learning techniques have been applied for churn prediction in the past decade. This section will cover some of these techniques and how well they performed when applied in the context of churn prediction.\par
\subsubsection{Support Vector Machines}
Support vector machines were first introduced by Vapnik during 1995 which were included in his studies in statistical learning theory. The main concept of SVM is to take known labelled data observations and map them in a linear feature space where the separation between the classes is maximized. This is done through an optimization algorithm which aims to maximize the separation margin between the classes \cite{boser}. Furthermore with the introduction of slack variables (usually denoted as C) in the optimization function, a “hard” or a “soft” margin can be achieved between classes. A “hard” margin will have a lower separation between the classes but will tend to misclassify less than a “soft” margin, which will be lenient towards misclassification but will allow for a larger separation between classes. In real world situations employing a “soft” margin might be preferable as not to overfit the model.\par
In practice however the data is not linearly separable. A way around this is to perform non- linear mapping of the input feature space into a high-dimensional feature space using of what is now popularly known as the “kernel trick” \cite{guyon}. This allows the support vector machine algorithm to generalize across different non-linearly separable data depending on the kernel function used.\par
The most recent application of Support Vector Machines in the context of churn prediction was identified to be used to predict churn in subscription services \cite{coussement}. The paper noted that the use of SVM’s was not well documented in published research and that previous implementations were based on unrealistic data with small sample sizes without much noise in their samples. Motivated by these reasons, SVM was applied to real data which was gathered from a subscription oriented Belgian newspaper, and its performance was compared to Logistic Regression and Random Forests techniques. Furthermore the paper noted that using SVMs has distinct advantages.\par
\begin{itemize}
	\item Support Vector Machines only require two parameters to be chosen in order for them to generate predictions. The kernel parameter and slack variable ‘C’.
	\item The model generated by SVMs is always optimal and global. This is extremely advantageous as other methods might fall into local minima during their parameter optimization.
\end{itemize}
The results of the approach presented, showed that SVMs perform very well in the application of churn prediction even on realistic, noisy datasets. The applied SVM was able to beat Logistic Regression but under performed when compared to Random Forests. Furthermore the paper notes that the performance of an SVM is greatly dependent on the parameters (kernel function and ‘C’) that it is given and in turn the parameters are depended on the data features. It was also noted that SVMs take significantly more time to train than Logistic Regression and Random Forests. This is definitely the biggest drawback of SVMs, as companies usually deal with very large, high-dimensional datasets. Even though the work done in the paper was a based on real data, SVM’s could prove to be unscalable in the world of big data.
\subsubsection{Decision Trees and Random Forests}
Decision tress have been used extensively in the context of churn prediction throughout the years. A decision tree can be thought of as a tree structure representation of a given classification or regression problem. It is composed of nodes which are also known as ‘non- leafs’ that represent explanatory variables. Subsequently a set of decisions to be made are ‘grown’ from these nodes, based on a subset of values of the explanatory variable the node depicts. This is repeated until the hierarchical representation generated has all of its end nodes linked to a value from the target variable \cite{muata}. This is more easily understood in Figure 2 from \cite{mitchell} which illustrates a decision tree grown from explanatory variables ‘Outlook’, ‘Humidity’ and ‘Wind’ in order to predict the categorical target variable ‘Play?’.
\begin{figure}
\begin{center}
	\includegraphics[width=3.5in]{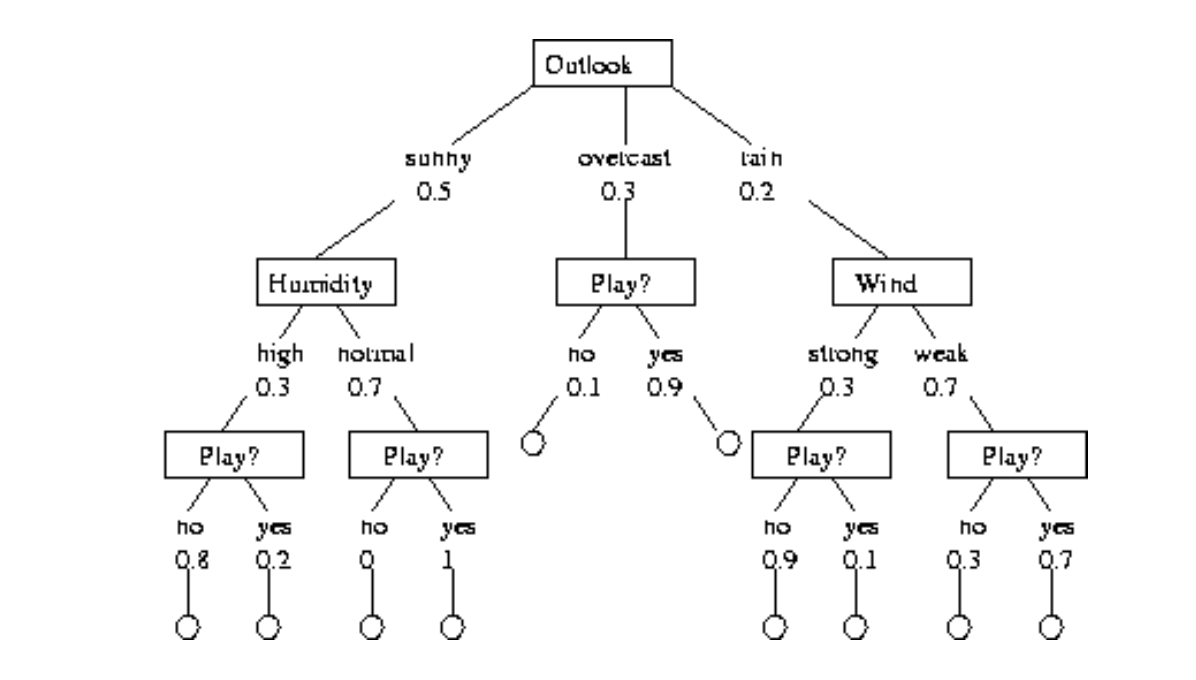}
	\caption{Example of a Decision Tree}
\end{center}
\end{figure}
A lot of algorithms have been developed in the last decade in order to build efficient and effective decision trees for machine learning applications (CART, C5.0 etc.) \cite{bloemer}. However single decision trees have proved to underperform when compared to other methods. Furthermore decision trees have a tendency to focus their growing on the majority class when presented with imbalanced datasets. Thus ensemble methods (usually bagging) were developed in order to address the poor performance of decision trees. This is performed by generating a lot of different decision trees that are able to work as a single classifier through majority voting on their predictions. One of the most popular ensemble classifier for decision trees is known as the Random Forests algorithm.\par
This is an ensemble classifier and thus has the distinct advantage of not over-fitting its generated models due to the Law of Large Numbers. The algorithm works by splitting the dataset into random subsets of samples and subsequently generating decision trees on each subset. During the prediction phase each tree is allowed to report its predictions and the majority prediction is the one returned by the model \cite{breiman}. By applying the right amount of randomness in their configurations, Random Forests can become extremely accurate classifiers. Furthermore due to their inherent process of creating multiple decision trees during the model generation, the algorithm is perfectly suited for being deployed in distributed systems (each node can build a distinct tree) which can dramatically decrease computation time in training and validation \cite{kulkanari}. Even though the effect of bagging allows the Random Forest algorithm to avoid overfitting, they still do not perform as well on datasets where there is extreme class imbalance; for example churn prediction datasets.\par
This inherent flaw is what motivated the development of Improved Balanced Random Forests \cite{xie}. The proposed algorithm combined two previous attempts on tackling this issue, Balanced Forests and Weighted Forests. Balanced Forests work by sub sampling a dataset while balancing the samples in terms of class distribution for each tree. This is repeated until all trees generated have covered the majority class. Weighted random forests assign weights to each class, such that the weight of the majority class has a lower weight than the minority class in order to penalize on misclassification accordingly.\par
The paper states that both these previous attempts have their limitations and continues by saying that the two previous attempts can be combined in order to make an extremely efficient and accurate classifier. The proposed algorithm was evaluated on real-world banking data provided by a Chinese bank and feature selection was done in order to select optimal features for the model. In order to compare the proposed algorithms’ performance, training and testing were performed on Artificial Neural Networks, Decision Trees, SVMs, and on both Balanced Forests and Weighted Forests. Results showed that the proposed algorithm outperformed both previous attempts as well as the other traditional approaches.\par
From the results it is easy for one to conclude that the Improved Balanced Random Forests is the state-of-the-art algorithm for churn prediction. In terms of predictive performance they outclass other methods, but also due to their effective scalability, fast training and fast predictive speeds they offer great potential in the problem of churn prediction. Having that said, their performance is dependent on the features selected, and therefore the feature engineering stage cannot be avoided when using this algorithm.\par
\subsection{Unsupervised Feature Learning}
It is argued that the only way to allow a machine to understand the world around it (AI) is by first being able to untangle hidden features from the data without needing a human to intervene. In order to address the issues that occur due to the effects of high- dimensionality, an unsupervised representation algorithm is required. The algorithm should be able to decompose complexities within datasets and consequently generate new, more effective features. This is what motivated the development of deep learning algorithms \cite{bengio:representation_learning,bengio:deep_architectures}.\par
The key findings that propelled the ingenuity behind the deep learning algorithms was the proposal of how the human brain lets visual information flow through a hierarchical neural network in its visual cortex in order to learn what is being observed by the patterns the information exhibits \cite{arel,bengio:deep_architectures}. Thus assuming that an algorithm could mimic this process, even at a very crude level, the algorithm could be applied over large datasets and consequently at every step of the hierarchy produce abstracted data features without the need for human supervision.\par
Since the introduction of deep learning algorithms, a number of different models have been created that have been able to simulate the effect that occurs in the human visual cortex using artificial neural networks. These models can be further generalized into three different types of deep architectures that have different types of applications \cite{deng:tutorial}:
\begin{itemize}
	\item \emph{Generative deep architectures} - used to describe the higher-level correlation properties of the observed data for pattern analysis, and consequently describe the joint statistical distributions of the observed data with their associated classes.
	\item \emph{Discriminative deep architectures} - used to directly classify patterns by describing previous distributions of classes given by the observed data.
	\item \emph{Hybrid deep architectures} - used for when the goal is to classify but is supported by the outcomes of a generative architecture. Usually these architectures have the highest prediction accuracy.
\end{itemize}
\par Most of these architectures are variants of other models and some are simply combinations of models especially in the case of hybrid architectures \cite{deng:tutorial}. The main concept is that they have some sort of hierarchy that is able to take inputs at the networks input layer and at every level in their hidden layer, create more abstract data features. This is done by having less artificial neurons at each step up the hierarchy \cite{bengio:deep_architectures}. Thus at the output layer of the network, extremely high level abstractions of the data are produced which are constructed by a high-resolution of previous data features. In the case of discriminative deep architectures better predictions can be made on classes based off these high level abstractions rather than the direct input features.

\subsection{Applications of Deep Learning in Churn Prediction}
Deep learning architectures have been successfully applied in various pattern recognition scenarios: image recognition, natural language processing and signal processing (mostly audio) \cite{le,deng:speech,deng:dlmethods}. Thus there should be no reason why deep learning could not be applied in churn prediction as it is simply the analysis of user behavioural patterns. Having that said, there are not a lot of scientific papers looking into the application of deep learning in churn prediction. In fact research was only able to identify one published paper describing this scenario, which discussed the application of deep learning in customer churn prediction regarding a mobile telecommunication network \cite{castanedo}.\par
The paper proposes a discriminative deep architecture using a four-layer feedforward neural network which acts as binary classifier which distinguishes churners and non- churners according to a users’ call patterns. The main motivation behind the use of deep learning architectures was to investigate the possibility of avoiding the time consuming feature engineering step in the company’s pipeline while at the same time beating their previous predicting performance.\par
Due to the high underlying complexities of user call interactions they needed to introduce a data representation architecture that could efficiently describe user behaviour across multiple layers while keeping the representation as detailed as possible. This was essential as deep architectures require a high-resolution input so that they can successfully unravel the underlying interactions and generate secondary features that can increase the separation between classes.\par
This data representation was used to train and test the deep feed-forward network with a sigmoid activation function in its hidden layers. Results have shown that the model is stable across most of the months which suggests that the model generalizes well and does not overfit the data. Furthermore the company was able to significantly increase their prediction accuracy from 73.2\% to 77.9\% AUC. Therefore it can be concluded that multi- layer feed forward models are effective in churn prediction.\par
The paper also noted that there are some possible enhancements and further research that can be attempted to further improve the models accuracy. Location data of calls could be included in the generation of their data representation architecture. Furthermore the paper hinted that Deep Belief Networks (generative architecture) could be applied as a pre-training step which could also increase performance of their multi-layer feedforward network.\par
The papers’ overall goal of using a deep learning architecture to avoid the feature engineering stage is very close to the goal of this project. However, the developed data representation is very domain specific (telecommunications industry) and moreover, it does not completely avoid feature engineering (a few features were engineered to compose the final segment of the data representation). Even though the papers’ secondary goal of beating the previous prediction accuracy was achieved, it did not prove that the feature engineering step could be avoided completely. Furthermore the data representation proposed cannot be applied to any company other than a telecommunications company.\par

\subsection{Diving Deeper into Deep Learning Mechanics}
This section covers in depth information regarding the inner mechanics of deep neural networks. It will provide context as to how deep architectures are able to learn, as well as the techniques that were used for implementing the final architecture proposed.\par
\subsubsection{Supervised Learning}
In the case of supervised learning on a specific training set $Z$, he goal is find a hypothesis function $f^h$ that approximates the function $f^*: X \rightarrow Y$, where $X$ is the feature set in $Z$, $Y$ is the output label (target variable) in $Z$ and $Z$ is of the form $\left \{ \left ( x_1, y_1 \right ), \cdots ,\left ( x_n, y_n \right ) \right \} \in \left ( X,Y \right )^n$ \cite{cunningham}. As all instances of $Y$ are known for all instances of $X$, it can be said that $Z$ is of the form $\left \{ \left ( x_1, f^*(x_1) \right ), \cdots ,\left ( x_n, f^*(x_n) \right ) \right \}$. Through the use of an appropriate cost function $J$, all the points of $Z$ can be used to find the parameters that fit $f^h$. Thus supervised learning consists of finding the minimum of the arguments of the cost function $J$, on training set $Z$:
$$f^h=argmin\left ( J\left ( Z \right ) \right )$$
\par It has to be noted that simply finding the approximation to the function which fits the training set is not enough for true supervised learning. The approximation computed needs to generalize well not to just the samples in the training set but also to new samples. Thus for optimized supervised learning, the function should be tested on subsequent validation and test sets in order to quantify its efficacy (how well it generalizes). \par
\subsubsection{Types of Artificial Neurons}
As mentioned previously the motivation behind the development of deep learning architectures was the hierarchical neural network in the human visual cortex \cite{arel,bengio:deep_architectures}. In order to simulate this, artificial neurons were used to form this hierarchy for computational simulations. Artificial neurons are simply computational units that take an arbitrary number of inputs (including a bias input) and through a specific activation function are able to return a single output. This is can be understood easier through the example shown in Figure 3.\par
\begin{figure}[!t]
\centering
\includegraphics[width=3.5in]{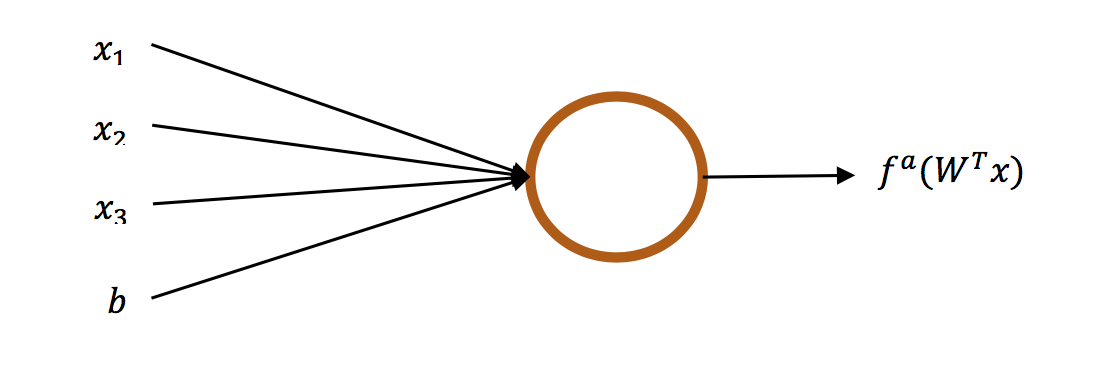}
\caption{Example of single artificial neuron}
\end{figure}
As shown in the figure, an artificial neuron takes inputs $x_1, x_2, x_3$ and a bias term $b$. In fact this is the simplest form of a neural network, which in this example allows for the representation of a hypothesis $h_{W,b} (x) = W_1x_1 + W_2x_2, + W_3x_3 + W_0b$. This can be generalized through an activation function $f^a:\Re\rightarrow\Re$ such that: $$f^a(W^Tx) = f^a\left ( \sum_{i=1}^{3} W_ix_i + b \right ) = f^a(z)$$
\par There are mainly two types of activation functions that are used in neural networks, a sigmoid function (logistic or hyperbolic tangent) or the more recently developed rectified linear function.\par
The use of sigmoid functions in deep neural networks stems from the fact that they introduce non-linearity to the model \cite{kocak}. The logistic sigmoid neuron generates a linear combination of its input values and weights (pre-activation z) and applies the logistic regression function to the result.
$$f(z) = \frac{1}{1+e^{-z}}, \text where: \space z =\sum_{i=1}^{k}W_ix_i + b$$
Thus the output of the neuron is bounded between 0 and 1. Intuitively the larger the value of the neurons pre-activation function, the closer the output will be to 1. Furthermore due to its easily calculated derivative $\frac{\mathrm{d}}{\mathrm{d} x}f(x)=f(x)(1-f(x))$, it allows the possibility for a network,
composed of these neurons, to be trained using greedy optimization learning algorithms like gradient descent. \par
Instead of the logistic function, a hyperbolic tangent (tanh) activation function can be used: $$f(z)=\tanh(z)=\frac{e^z-e^{-z}}{e^z+e^{-z}}, where: z=\sum_{i=1}^{k}W_ix_i + b$$
This is simply a rescaled logistic activation function with its lower limit set to -1. Studies have shown that when presented with normalized data (between 0 and 1), hyperbolic tangent activation functions seem to generate stronger gradients during backpropagation (data is cantered on 0 thus derivatives are larger) \cite{lecun:backprop}. \par
The rectified linear activation function has no upper limit above 0 and any negative pre-activation value computed will be set to 0.
$$f(z)=max(0,z), where: z=\sum_{i-1}^{k}W_ix_i + b$$The effects of this is that a rectified linear function can only have two possible derivatives and thus the output values can only be 0 or $W_i$ \cite{glorot}.This makes rectified linear neurons extremely computationally efficient. Moreover the general effect of employing rectified linear neurons in a network, is the fact that they allow the network to form sparse propagation paths as neurons will either be active or not and thus computations become linear along these paths. Due to this linearity, gradients do not “vanish” during backpropagation as can be noticed in sigmoid or $tanh$ activation functions.\par
\subsubsection{Deep Feed-Forward Neural Network (Multilayer Perceptron}
In theory stacking artificial neurons in various different combinations can allow such computational units to solve ever more complicated functions. More formally any function can be represented by a set of computational units configured in certain way \cite{bengio:deep_architectures}.
\begin{figure}[!t]
\centering
\includegraphics[width=3.5in]{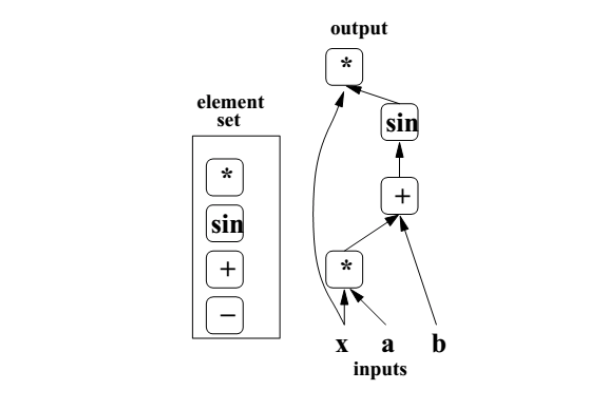}
\caption{Graph of a function computing $\protect{f(x)=xsin(ax +b)}$}
\end{figure}
The configuration and connections between elements can be represented by a graph. For example the expression of the function $f(x)=xsin(ax +b)$ can be considered as the composition of a set of operations which is illustrated in Figure 4 (from \cite{bengio:deep_architectures}). \par
Through this intuitive example one can recognize that a complex function, such as $f(x)$ cannot be expressed through a single computational unit of a specific type. This is similar to how complex non-linear functions can be approximated by stacking artificial neurons and by subsequently training each neurons weights. A feed-forward neural network or MLP is made up of an arrangement of interconnected neurons with a simple activation function. The arrangement of an MLP can be seen in Figure 5 (from \cite{dorling}) and simulates the non-linear mapping of an input vector to an output value.\par
\begin{figure}[!t]
\centering
\includegraphics[width=3.5in]{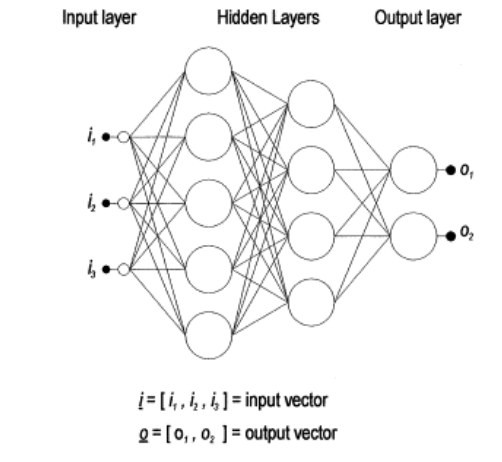}
\caption{Graph of a Multilayer Perceptron}
\end{figure}
Initially an MLP has no inherent ability to solve a highly complex non-linear function as its weights are initialized randomly upon instantiation. Thus by allowing each layer of neurons to propagate their activations to forward layers (feed-forward architecture) and iteratively ‘fix’ their weights during backpropagation, the composition of weights of the architecture will begin to give increasingly better approximations to any function in every iteration. Furthermore, after successful training, each neuron in the network can be thought of as a feature detector \cite{bengio:deep_architectures}. Thus by allowing information to flow forward between hidden layers, more complex, abstracted features will be generated within the network. This is because top level hidden layer neurons assign their weights based on the activations of previous hidden layers, which in theory are a combination of complex feature detections.\par
In the case of classification, a $softmax$ function can be used as the activation function in each artificial neuron located in the output layer. This is a generalization of logistic regression so that multiple classes can be predicted by having a neuron for each class that exists (can also be used for binary classification by having 2 neurons) \cite{ng:tutorial}. $$P\left ( y^{(i)} = k | x^{(i)}; W; b \right ) = \frac{exp\left ( W^{(k)T}x^{(i)}+b^{(k)} \right )}{\sum_{j=1}^{k} exp\left ( W^{(j)T}x^{(i)}+b^{(j)} \right )}$$
Each neuron in the softmax layer will make a prediction $y^{(i)}$ based on its inputs $x^{(i)}$, its weights $W$ and its bias value $b$. The $softmax$ layer differs from other layers in an MLP as its neurons work collectively to put into effect $\sum_{i=1}^{k}y^{(i)}=1$. In other words each neuron will return a probability as to how likely the propagated inputs belong to its class, such that all the probabilities returned sum up to 1. Thus the predicted class can be found by the position of the neuron in the output layer which returned the highest probability.$$y_{predicted} = argmax\left ( P\left ( y^{(i)} = k|x^{(i)}; W; b \right ) \right )$$ Consequently the final prediction may not be equal to the actual value of the target variable. The difference between the output predictions and the actual outputs can be quantified as an error signal, or otherwise the cost \cite{dorling}. The magnitude of the cost is what determines how the weights will be adjusted during backpropagation. The regularized cost function $J(W)$ for an MLP demonstrating classification through $softmax$ is given by the following function
	\cite{ng:tutorial}.
\begin{equation}
	\nonumber
	\resizebox{3.5in}{!}{$J(W) =\frac{1}{m}\left [ \sum_{i=1}^{m}\sum_{j=1}^{k}1\left \{ y^{(i)} = j\right \}\log \frac{e^{W_{j}^{T}x^{(i)}}}{\sum_{l=1}^{k} e^{W_{l}^{T}x^{(i)}}}\right ] + \frac{\lambda}{2}\sum_{i=1}^{k}\sum_{j=0}^{n}W_{i,j}^2$}
\end{equation}
This is equivalent to the negative log likelihood with an added “weight decay” regularization technique which essentially limits overfitting of the weight parameters (L2 regularization). From the above function we can see that due to the indicator function $1\left\{ y^{(i)}=j\right\}$, only the activation neuron at position $j$ will contribute to the cost. Intuitively by looking at the plot of $-\log(x)$, one can recognize that as the activation of a particular neuron gets closer to 0, the cost increases exponentially. Thus by forcing the model to minimize this error would essentially force the weights of the network to promote an activation close to 1 for that particular neuron. \par
By varying the weights across all possible values and passing it through the cost function, an error surface could be generated \cite{dorling}. It is difficult to visualize this for all the weights as a plot of all the weights in a normal MLP would most likely exceed three dimensions. In order to give context to gradient descent and backpropagation, an error surface visualization for an MLP with only two weights is shown in Figure 6 (from \cite{dorling}).
\begin{figure}[!t]
\centering
\includegraphics[width=3.5in]{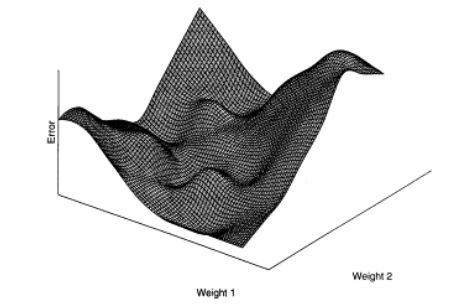}
\caption{Error surface generated through cost function between two weights}
\end{figure}
The error surface depicted from the two weights shows how the error changes with respect to the values of the weights. Essentially the absolute goal of a backpropagation algorithm is to locate the global minimum of the errors surface. It is able to do this through a technique known as gradient descent.\par
As mentioned previously the weights of an MLP are instantiated randomly which technically means that that a random point in the errors surface is selected. In order for it to decide on how to alter the values of the two weights, the gradient of that locally selected point is calculated. This is done by differentiating the cost function with respect to each weight. By taking derivatives it can be shown that the gradient can be calculated using the following function \cite{ng:tutorial}.
\begin{equation}
	\nonumber
	\resizebox{3.5in}{!}{$\bigtriangledown_{W_{j}}J(W) = -\frac{1}{m}\sum_{i=1}^{m}\left [ x^{(i)}\left ( 1\left \{ y^{(i)} =j \right \} - p\left ( y^{(i)}=j|x^{(i)};W \right ) \right ) \right ] + \lambda W_j$}
\end{equation}
Thus by using the resultant partial derivative vector $\bigtriangledown_{W_{j}}J(W)$, the weights can be updated according to their partial derivative and a constant $a$, which is more formally known as the learning rate.$$W_j:=W_j - a\bigtriangledown_{W_{j}}J(W)$$ The whole process is repeated and the weights are updated iteratively until the backpropagation algorithm is satisfied that it has reached the global minimum of the error surface. Of course almost all backpropagation algorithms that use gradient descent in order to optimize a networks weights, cannot be certain that they have reached a global minimum. As can be seen in Figure 6, the error surface is composed of multiple local minima and thus a simple backpropagation algorithm could get stuck in a suboptimal local minimum. This is where the learning rate plays a big role in assisting the backpropagation algorithm to determine if it has reached a reasonable error minimum. \par
The learning rate can be thought of as the ‘step-size’ the algorithm takes as it goes down an error slope. A very large learning rate will cause the backpropagation algorithm to repetitively miss a global minimum due to the erratic weight changes. Similarly if the learning rate is too small, backpropagation will be very slow and may never reach the best local minimum, as it might get caught in a different local minimum. To address this issue the momentum parameter can be introduced in the backpropagation algorithm which can help “push” the descent over local suboptimal minima \cite{qian}. It does this by incorporating a proportion of the previous weight update in the current weight update. Thus the weight update would happen as follows: $$W_j:=W_j - a\bigtriangledown_{W_{j}}J(W) + p\bigtriangleup W_{j-1}$$ \par
The backpropagation algorithm described above is known as the batch backpropagation algorithm as for every iteration in its process, all training samples are used to generate the gradients, which in turn update the weights \cite{ng:tutorial}. Studies have shown that in order to get more accurate weight updates, the gradients are computed over “mini-batches” (subsets) of the complete training set \cite{bengio:deep_architectures}. Subsequently the average of these gradients is taken to be the final update to be performed on the weights. This is more formally known as Stochastic Gradient Descent (SGD) or as mini-batch gradient descent.
\subsubsection{Tackling Overfitting in Deep Neural Networks}
In supervised learning applications of deep neural networks, as in the architecture discussed above, optimal predictive performance is not a measure of how well the architecture can fit the given output values (training set). Success in any supervised learning application is quantified as to how well the model performs on data it has not been trained on. Due to the fact that deep learning architectures generate more complex, abstract features (usually higher order features) in each hidden layer, they tend to overfit the training data. This can be demonstrated through the following example of a linear model versus a high order polynomial model in Figure 7 (from \cite{buduma}).
\begin{figure}[!t]
\centering
\includegraphics[width=3.5in]{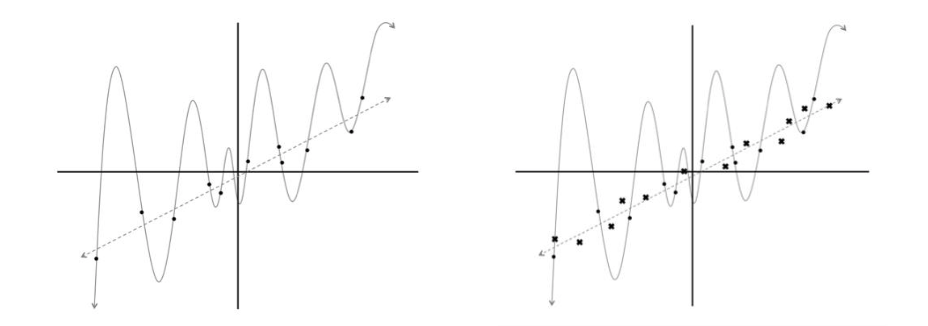}
\caption{Comparison of linear model vs high order polynomial model: (Left) Training Data, (Right) Training Data with Test Data}
\end{figure}
As can be seen in Figure 7, a linear and a polynomial model were fit on the initial training data (left plot). One could come to the conclusion that the polynomial model fits the data perfectly and thus is the “better” model. However, when presented with additional data from a test set (right plot), the polynomial model performs significantly worse than the linear model which in fact was the best suited model or in other words was the model that best generalized the variances within the data. In order to address this intrinsic issue of deep learning architectures, regularization parameters are introduced within a cost function so that large computed weights are penalized. The most recognized regularization parameters that are commonly used in deep neural networks are the L1 regularization and the L2 regularization \cite{ng:regularization}. \par $$L1:\lambda\sum_{i=1}^{k}\left | W_i \right |$$
L1 regularization is basically the sum of the weights multiplied by an L1 constant $\lambda$. This type of regularization has the interesting effect of making the weight vector very sparse. In other words most of the weights will be close to zero with only a few large weights. This is important as it forces neurons to compute their activations on the most important inputs, making them more resistant to noise. \par $$L2: \frac{\lambda}{2}\sum_{i=1}^{k}W_{i}^{2}$$
L2 regularization or “weight decay” is the sum of the square of the weights multiplied by half the L2 constant $\lambda$. The regularizer has the intuitive effect of forcing the cost function to penalize high variance weight vectors, thus forcing the selection of weight vectors with less variance in their weight values. By doing this, the network is forced to use all of its inputs, rather than using specific inputs over and over. \par
Additionally to these regularization techniques, an early stopping mechanic can be implemented within a backpropagation algorithm. This mechanic controls the overall run time of the algorithm. In order to be able to apply it, the data needs to be split into training, validation and test data. Its main purpose is to allow training to continue on the training data as long as the validation error decreases \cite{prechelt}. This ensures that the architecture does not begin to overfit the training data. Furthermore this allows the training time to be optimal where further training would not produce any significant increase in prediction accuracy. \par
Another very recent technique for preventing overfitting is Dropout \cite{srivastava}. The technique works by randomly dropping (turning off) a proportion of neurons from a specific layer in every training iteration. It does this by giving each neuron in a layer a probability $p$ that it will be active. Applying Dropout during training has the effect of running training on a subsample of the actual network. This is because any neural networks of $n$ artificial neurons can be seen as a group of $2^n$ possible subsamples. Thus throughout the whole training process of a network with dropout, can be thought of as the training of $2^n$ subsamples. At test time the average prediction of all subsamples is taken. This brings distinct advancements to any deep learning architecture as the initial architecture inherits advantages seen in ensemble machine learning algorithms. This includes the generation of a much more generalized non-linear function with less overfitting on unseen data.
\subsection{Scaling the Generation of the Encompassing Data Representation Architecture}
Due to the extremely massive amounts of data that Framed handles through company event logs, generation of the encompassing data representation architecture on a single machine proved to be impossible. Thus alternative technologies were investigated in order to address this issue. This section will cover the technologies used in order to allow the generation of the data representation to be realized.
\subsubsection{Apache Spark Cluster Computing Engine}
Cluster computing refers to the process of performing parallel data computations on a cluster of computers. This is done through the use of a system that can defragment the complexities of a single computation and assign fragments of the computation to be performed on different computers in the cluster \cite{valentini}.Thus the overall completion of the main computation can be collected from each computer and combined as a single object. This model was realized through the now extremely popular MapReduce \cite{dean} system. The system provides a user with a programming model where he/she can pass data through a set of operations in a created acyclic data flow graph. An example of a directed acyclic graph (DAG) can be seen in Figure 8. Therefore a complex computation can be thought of as a set of simple operations that have to be completed in a specific order so that the result satisfies the result of the original computation. The order of operations of the example in Figure 8 would be \{5, 3, 1, 4, 2\}. \par
\begin{figure}[!t]
\centering
\includegraphics[width=3.5in]{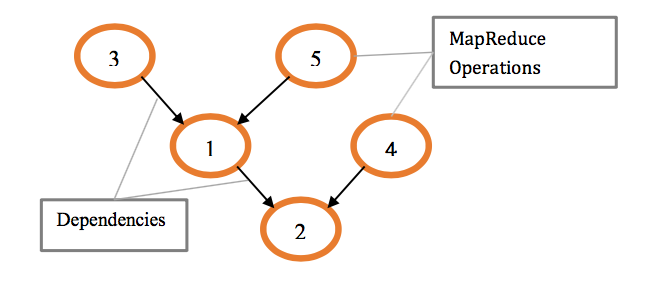}
\caption{Directed Acyclic Graph of operations to be applied so that the overall computation can be completed}
\end{figure}
Even though this data flow programming model can be applied to satisfy a lot of different applications, applications that reuse intermediate results of the computation in a lot of parallel operations, cannot be efficiently expressed through such a model \cite{zahariaetal}. An example of such an application is an iterative application where each iteration is expressed as set of operations to be performed. This inherently would mean that each iteration would invoke data reloading from storage, which would significantly decrease computation performance. Spark is a relatively new cluster computing framework that tackles this issue through its novel abstracted programming model of the two stage MapReduce model. This novel programming model allows spark to support any arbitrary acyclic graph of operations.\par
Spark introduces the concept of resilient distributed datasets (RDDs) \cite{zaharia:rdd} which are a representation of the effect of certain operations/transformations on data in storage or in other RDDs. This means that Spark applications can be written in a sequential format through a chain of RDDs which is a much more intuitive way for a programmer to describe the flow of data operations; and in effect be able to reuse previous RDDs in iterative computations. As RDDs can be cached to memory, reusing a single RDD (stored in memory) for iterative operations, allows Spark to be 100x faster than Hadoop MapReduce. Furthermore these transformations are applied “lazily” such that they can be stacked in a sequence without any computation occurring in the background. Essentially through a stack of transformations, a directed acyclic graph is generated in the background.\par
Only the application of an “action” such as $reduce$ or $collect$ will trigger the initialization of a computation. After a computation is initialized, the graph is passed to the DAGScheduler where the graph is split into stages of tasks. These stages are comprised of a set of optimized operations to be performed on the data. This optimization ability of the DAGScheduler is what grants Spark the ability to be 10x faster than Hadoop MapReduce on disk operations. For example, multiple $map$ transformations can be scheduled within a single stage.\par
Once all stages are optimized, they are flagged as ready and subsequently passed to the TaskScheduler which initiates tasks via the employed cluster manager which could be simply Spark Standalone or even Yarn or Mesos. Each task in a stage is executed by a worker in the cluster and its computation results (blocks) are stored in a workers’ memory, which in turn can be returned for computations on subsequent stages. It has to be mentioned that the TaskScheduler is not aware of any stage dependencies. Thus stages to be computed that have dependencies on previous completed stages, whose task results have been discarded from memory, are recomputed \cite{zahariaetal}. This allows Spark applications to be resilient to faults but comes as a cost of reduced performance if there isn’t enough cluster memory to store all task results.\par
\subsubsection{Hadoop Distributed File System}
In order for all of Spark workers to have access to the data so that they can perform their individual tasks, the data needs to be distributed. Furthermore workers need to have access to data that might be stored on different workers. This can be achieved through the implementation of a distributed file system such as the Hadoop Distributed File System (HDFS) \cite{shvachko}.\par
HDFS allows for highly scalable distributed storage of data and it is the basis for all Hadoop applications. It is able to do this by separating file system metadata and application data into NameNode and DataNode servers respectively. File system metadata take the form of $inode$ objects which contain file and directory information such as permissions, modification times and access times. File content is split into blocks which are large size chunks of a files data. The blocks are subsequently distributed and reproduced in a number of DataNode servers (most commonly three). This can be seen in Figure 9 which depicts an HDFS architecture with client interactions.
\begin{figure}[!t]
\centering
\includegraphics[width=3.5in]{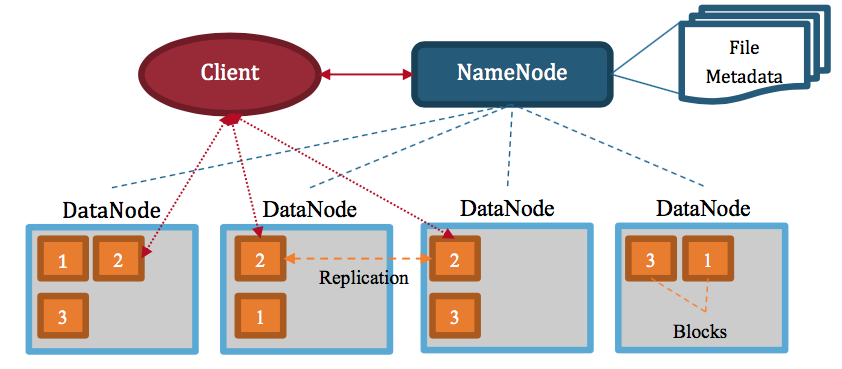}
\caption{HDFS Architecture with client interactions}
\end{figure}
The diagram above assumes that file data can be held in single blocks (block numbering). This was done in the hopes of demonstrating the effect of block replication. In reality a single data file will be split into multiple blocks depending on the chunk size selected and the size of the actual data file. \par
Furthermore the figure demonstrates how a client interacts with the HDFS architecture. Regardless of whether the client wishes to write or read data, client interactions are initiated on the NameNode server where file metadata and namespace tree information are held. If a client wants to read particular file’s data, the NameNode server is contacted and the locations of the blocks regarding that particular file are returned by the NameNode. Subsequently the client then reads the block data directly from the DataNodes that the blocks are stored in. Similarly if a client wants to write data to the HDFS, the NameNode is contacted with a request of nominating three suitable DataNodes where file data blocks can be stored and replicated. After the NameNode returns this information from its namespace tree, the client then proceeds to writing and replicating the blocks directly on the three DataNode servers in a sequential fashion.\par
In order to keep the overall system integrity, DataNodes send $heartbeats$ to the NameNode which contain information about a DataNodes’ status and the blocks hosted on that DataNode. This is usually done every three seconds and a DataNode failing to do so for ten minutes will be regarded as out of service by the NameNode. This in turn will initiate block replications of the blocks contained in the faulty DataNode on other “alive” DataNodes. Additionally heartbeats play an important role in order for a NameNode to perform efficient space allocation tasks and load balancing decisions. These actions are performed as response to heartbeats as a NameNode will not directly contact a DataNode. Thus it is critical that heartbeats from DataNodes to a NameNode are performed as frequently as possible.
\section{Methodology}
This section will cover the steps undertaken in implementing the previously described objectives. It will describe the reasoning behind the proposed data representation architecture as well as how it was realized through the use of a Spark computation cluster. Furthermore this section will cover the development progression of the proposed deep neural network through the addition of the more advanced deep learning mechanics described in the background research section.
\subsection{Encompassing Data Representation Architecture}
Each company that Framed deals with, tracks system events and is able to log them as JSON objects. Consequently daily event logs are supplied to Framed as “raw-dumps”, which are made up of these JSON objects, such that each JSON object is separated by a new line. An example of the general structure of a JSON object can be seen in Figure 10.
\begin{figure}[!t]
\centering
\includegraphics[width=3.5in]{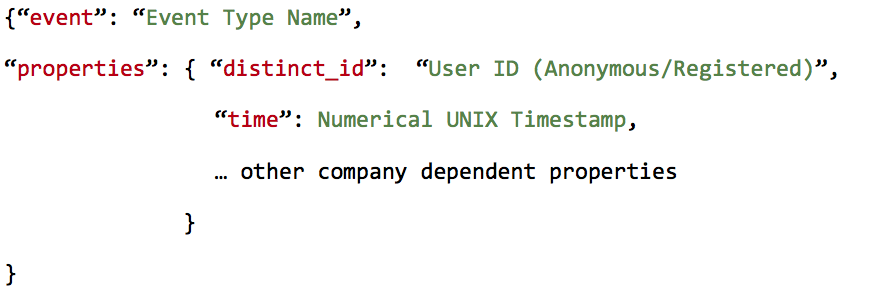}
\caption{Example of an event JSON object}
\end{figure}
As can be seen in the example, the \emph{“event”} key in the JSON object exists in every event logged and is independent of company type. In other words, any event from any company, will always contain this initial key in their JSON event objects. Even though the \emph{“properties”} key also exists in all JSON event object regardless of company, the value (object) of the \emph{“properties”} key changes for different types of events and also different companies will have different property value objects which depend on their system.\par
Since the data representation to be developed is concerned with user event data only, a key in the \emph{“properties”} value object needed to be identified that would indicate this. This was identified to be the \emph{“distinct\_id”} key. The \emph{“distinct\_id”} is only present in the \emph{“properties”} value object if the event logged was triggered by a user in the system. Furthermore the value of this key is independent of whether the user is a registered user in the company or a general/anonymous user. Having that said, there is a distinct difference between the values of registered and general user. Registered users have numerical distinct ids, while general users have long alphanumeric ids usually corresponding to system cookie ids. Lastly the \emph{“time”} key in the \emph{“properties”} value object was found to exist in all JSON event objects regardless of event type and company. The value of the \emph{“time”} key contains a numerical UNIX timestamp. All other keys in the “properties” value object were found to be company and event type specific.\par
By having realized what information was available in user event data across different companies, it was decided that the encompassing data representation architecture needed to be formed from these persisting key-value pairs. Thus it was essential that the values of the “event”, \emph{“distinct\_id”} and \emph{“time”} keys were scrapped from the daily \emph{“raw-dump”} files in a reasonable data structure. The values from these keys were scrapped from JSON event objects and stored in tuples of the form $\{userid, time, event\}$. After collecting all the tuples formed from each JSON event object, tuples containing user ids of non-registered users needed to be removed from the collection. This was done by validating that the \emph{userid} was completely numerical, as it was known that non-registered users would have long alphanumeric values.\par
Having gathered the selected values, the question at hand was how these values could be structured in a way as to express differences between user behavioural patterns. Inspired by the representation used to mine user development signals in online community platforms \cite{rowe}, it was decided a user event vector needed to be generated for each user across a specific timeframe. Figure 11 depicts the proposed structure of a users’ event vector for a specific time frame.
\begin{figure}[!t]
\centering
\includegraphics[width=3.5in]{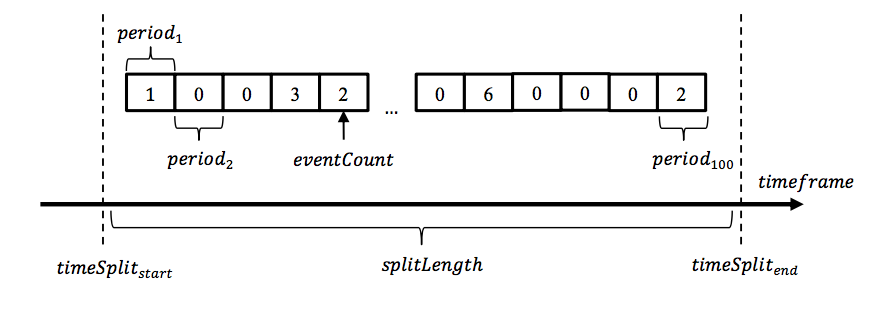}
\caption{Proposed user event vector structure for a specific user}
\end{figure}
In the context of churn prediction, user behavioural patterns need to analysed in order to predict whether based on those patterns, the user will churn or not. Since Framed provides results of this analysis on a monthly basis, the generated user event vectors needed to be confined within a specific $split$ of the complete input data timeframe. The timeframe of a split would be further subdivided into one hundred periods. The periods could intuitively be thought of as percentage positions of a splits’ time interval. Based on those period time intervals, user vectors could be generated for each user with one hundred dimensions. Each dimension in the vector represents a count of events that occurred in a period of that splits timeframe by a specific user. This is repeated for each customer so that after collecting all the user vectors, the end result would be a matrix of $N \times 100$ where $N$ represents the number of users. The heat map in Figure 12 depicts such a matrix.\par
\begin{figure}[!t]
\centering
\includegraphics[width=3.5in]{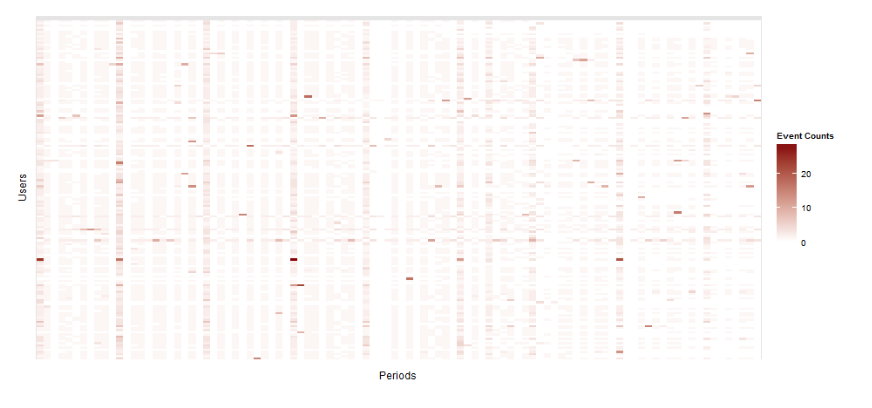}
\caption{Collection of user event vectors as a matrix}
\end{figure}
Even though the proposed representation is using very simplistic features to form user event vectors, intuitively this representation is able to capture the differences between user behaviours as event counts are essentially compared on percentages of time across users. As can be seen in the above figure, the generated user event vectors are sparse, but by having the split length parametrized, the sparsity of the vectors can be adjusted, which ultimately removes sparsity in the vector. Of course this is greatly dependent on the rate of user events of a particular company. Event vectors, with adequate sparsity, can be generated from a company with a very high rate of events, by using a small split length. Similarly extending the split length can benefit user event vectors of a company with not a lot of user activity. Thus it can be said that the split length parameter needs to be selected through trial and error, so that denser user event vectors can be generated with more pattern information. \par
Now that the proposed representation has been defined, the following step was to decide how to label these representations. Through their experience, Framed has realized that almost 81\% of the companies they deal with have no event implemented to signify the churning of a user \cite{framed}. Thus output values needed to be generated through some kind of logic that would determine if a user has churned based on the given data. Using the logic described in \cite{rowe}, a user would be deemed as a churner if there were no events triggered for a specific number of consecutive days. Due to the inherent business model of subscription companies to bill their customers on a monthly basis, it was decided that 30 days of inactivity would be a good threshold to signify user churn. This is because a whole month of inactivity means a loss in revenue for these companies, therefore a churner would have to be identified before they become inactive.\par
The concept of time split's used to generate the user event input vectors was also used to implement this logic. Since data from a company was divided into splits of a selected split length, data from subsequent splits could be used to determine the output value of a previous splits’ user event vector. This logic can be better demonstrated through Figure 13 which shows the complete development of training and validation/test sets.\par
\begin{figure}[!t]
\centering
\includegraphics[width=3.5in]{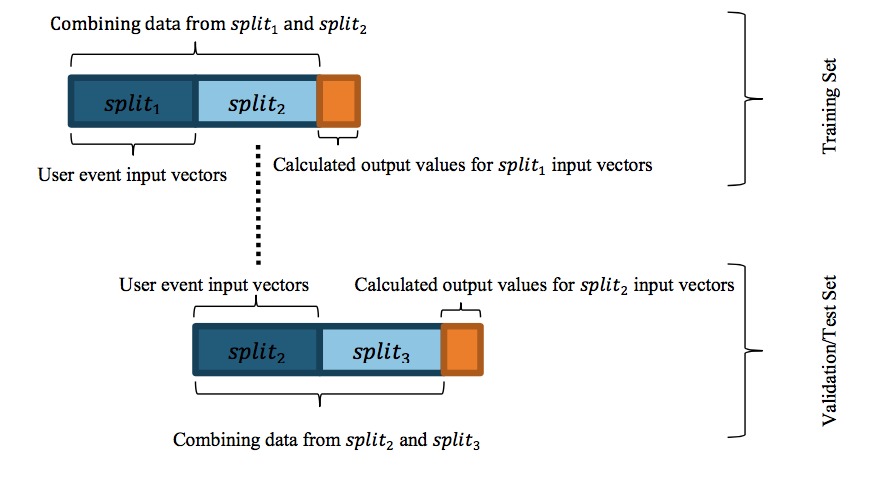}
\caption{Generating Training and Validation/Test sets using split data to define output values}
\end{figure}
The above figure shows the overall logic of using the proposed representation architecture to generate training, validation and test sets across 90 days’ worth of company data for a set split length of 30 days (split length can be varied). Let $split_1, split_2, split_3$ indicate consecutive 30 day splits. Initially $split_1$ data is used to generate user event vectors for each user present within the time interval of $split_1$. Then the combined data from $split_1$ and $split_2$ is used to compute churn output values for every user in the combined time interval of $split_1$ and $split_2$. The output values are calculated based on the time difference between the time of a users’ last triggered event and the final time of $split_2$. If this time difference is greater than 30 days, then a 1 is returned as an output value (indicating churn), otherwise a 0 is returned (indicating an active user). Furthermore, in the case that a users’ overall event time span throughout $split_1$ and $split_2$ is found to be less than 30 days (which would indicate a new user), a -1 is returned. This is important as a company would not be interested in keeping new users but rather retain long active users. Therefore any users with a value of -1 are filtered out.\par
The user event vectors generated for $split_1$ are joined with the generated output values based on their user ids. This is done as we are only interested in the output values of users in the time interval of $split_1$ and thus any other users that have registered after $split_1$ are not included. Therefore the end result will be a complete dataset for $split_1$ with user event input vectors and churn output values. \par
Since the overall aim of generating these representations is to effectively train deep neural networks to predict churn rather than just identifying what input vectors indicate churn, the deep neural network should be validated and tested on a following splits’ dataset. This is also demonstrated in Figure 13, as the whole process of generating a dataset for $split_1$ is repeated for $split_2$, using $split_3$ to generate its output values. Therefore the final result will be a dataset for $split_1$ and a dataset for $split_2$. The dataset generated for $split_1$ would be used to train a deep neural network while the dataset generated for $split_2$ would be further sub divided into validation and test sets (50\% - 50\% random split). Thus this ultimately would force the deep neural network to find a function that could predict churn on $split_2$ data representations based on $split_1$ data representations. This is key in enforcing prediction rather than just identification of data representations. \par
Due to the inherent nature of the problem of churn prediction, generated datasets will be imbalanced in terms of samples available for their respective output value classes. This is because customer churn will usually be a rare event. This can cause serious issues in prediction performance \cite{burez}, as a model will adjust its parameters to fit the majority class while disregarding the minority class. In order to address this issue, generated training, validation and test sets were balanced using random under-sampling. This method randomly removes samples from the majority class until the samples of each class are balanced. Naturally this may cause the loss of a considerable amount of majority samples that can contribute to better separation between the two classes, but this was the only identified technique that would not cause model overfitting.\par
\subsection{Scaling Dataset Generation with Spark}
Initially the generation of datasets was attempted using Python and various SciPy ecosystem packages such as NumPy and Pandas. Even after boosting the performance of certain iterations in the developed script by using multithreading Python techniques, dataset generation based on the proposed data representation architecture, could not be realized on a single machine. This was mainly due to the massive sizes of the raw-dump daily JSON event files which in turn caused extremely long iteration times. Therefore other technologies needed to be investigated that could generate these datasets as fast as possible and irrespective of how many days of data were selected. It was decided that a Spark computational cluster would be used as the literature stated that it could be up to 100x faster than a Hadoop MapReduce system (if there is enough system memory).\par
Framed graciously provided access to a Compute Engine project on the Google Cloud Platform in order build the Spark cluster. Compute Engine projects allow a user to create high performance virtual machines of various computational and memory specifications. By initializing a number of such virtual machines and by consequently installing Spark and Hadoop (HDFS) proprietary software on each one, they could be configured to work together as a computational cluster.\par
Before any building of the cluster could commence, the general architecture of the cluster needed to be considered. Since a Compute Engine project will only allow a maximum of 24 computational cores to be utilized across all virtual machines, it was decided that the Spark driver (master) would be based on a virtual machine with 8 cores and 16 Spark workers (slaves) will be based on single core virtual machines. This setup would allow for maximum computational performance while also allowing for a very powerful master server to perform any non-cluster data operations.\par
Furthermore it was decided that an HDFS architecture should be incorporated with Sparks’ architecture so that no additional virtual machines would need to be created. This was done by having the master virtual machine be both a Spark driver and an HDFS NameNode. Thus the remaining 16 slave virtual machines would also serve a dual purpose, as they would be both a Spark worker and an HDFS DataNode. The final architecture can be visualized in Figure 14.\par
\begin{figure}[!t]
\centering
\includegraphics[width=3.5in]{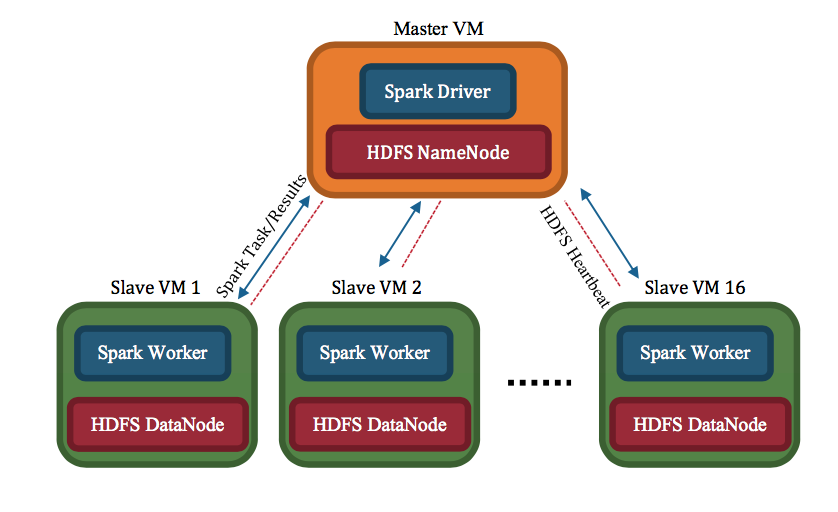}
\caption{Diagram of final Spark computational cluster architecture}
\end{figure}
Before any virtual machine instantiation, a virtual network needed to be created that would allow all the machines to operate under the same IP range. Furthermore this was important as the computational cluster needed special firewall entries to allow TCP, UDP and ICMP protocols to be used for communication between them. Thus by creating a virtual network these firewall rules could be applied internally within the Compute Engine without any concerns for outside security threats that could arise by enabling them.\par
Therefore each virtual machine instance was created on this virtual network following the specifications described in the architecture. It was important that all instances were working on the same operating system so that software installations could be carried out the same way across all machines. The chosen operating system was Ubuntu 14.04 LTS as it was a stable release of the popular Linux distribution. After all the virtual machines were instantiated, Java and the Java development kit were installed on the all the machines, as both Spark and Hadoop HDFS require Java to operate.\par
Spark allows for the possibility to operate in Standalone mode. This means that it does not require the installation of third party cluster managers (YARN or Mesos) in order for it to function. This can be achieved by installing a compiled version of Spark on each machine. Spark installations provide scripts that can be run in order to configure a cluster. A master server can be launched using the \emph{“start-master.sh”} script and a slave server can be assigned to a master server using the \emph{“start-slave.sh”} script followed by the IP of the master server. This process can be accelerated by adding the IPs of all slave servers to a \emph{“slaves”} file in the master servers’ configuration and consequently running the \emph{“start-all.sh”} script on the master server. In order for any of these scripts to work, password-less secure shell (SSH) access needed to be established between a master server and all slave servers. This was done by generating private and public SSH keys on the master server and by sequentially transferring these SSH keys to all slave servers. This allowed two-way communication between the master server and slave servers. Finally, after further configuration in the master servers’ Spark configuration files regarding environment variables, Spark was initialized using the \emph{“start-all.sh”} script on the master server.\par
To extend the capacity of the proposed architecture, additional 500GB drives were attached on each of the slave virtual machines. This was done using the Google Cloud SDK which allows for quick access and control of all projects in the Google Cloud Platform with simple shell commands. After attaching the drives to the slave virtual machines, the drives were mounted on each slaves’ operating system in identical directories. The directories would be used to hold the blocks of split data files within the HDFS.\par
Having everything set up, the next stage was to tailor the proposed data representations script so that it can utilize the Spark computational cluster. Essentially the script was re- written using the PySpark API which exposes the Spark programming model to Python. In other words RDD transformations could be executed through this API by supplying Python functions to the transformation methods, which would subsequently return Python collection types if a Spark “action” is executed on a transformation. Thus most of the scripts logic was split into functions that could be easily passed into RDD transformations, by following the respective transformation arguments and return prerequisites. \par
\begin{figure}[!t]
\centering
\includegraphics[width=3.5in]{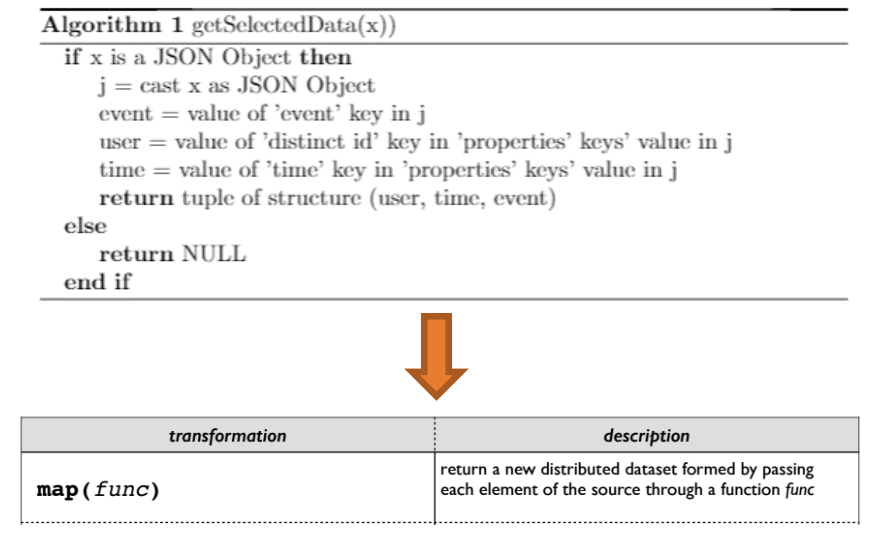}
\caption{Example of a function that can be passed through a map Spark transformation to gather the required data from a “raw-dump” file loaded in an RDD}
\end{figure}
The example shown in Figure 15, depicts an algorithm that can be implemented in Python and subsequently passed into a Spark $map$ transformation. After a “raw-dump” files’ data has been loaded as text elements in an RDD, the “getSelectedData” function can be performed through the $map$ transformation. This will return a new RDD where every text element in the original RDD, has been transformed into a tuple of $(userid, time, event)$. This is in fact is the first step discussed in the encompassing data representations’ architecture, where required information was scrapped JSON event objects.\par
The above methodology was applied to all procedures required to generate datasets of the discussed representation architecture. Intuitively a directed acyclic graph could be visualized from the performed transformations which would give better context to the overall way the Spark script works. Figure 16, illustrates how all the procedures needed to retrieve Event Vector to Churn Output mappings of the proposed data representation architecture, can be performed through Spark transformations in a directed acyclic graph. Transformations are illustrated as ovals and actions are illustrated as rectangles. The mapping for $split_1$ is consequently used to generate the training set, while the mapping for $split_2$ will be further subdivided to generate validation and test sets. Of course further procedures are performed in order to generate the final dataset from the two resulting mappings (like dataset balancing), but due to the fact that most of the processing is performed on the Spark cluster the complete process is extremely fast.\par
\begin{figure}[!t]
\centering
\includegraphics[width=3.5in]{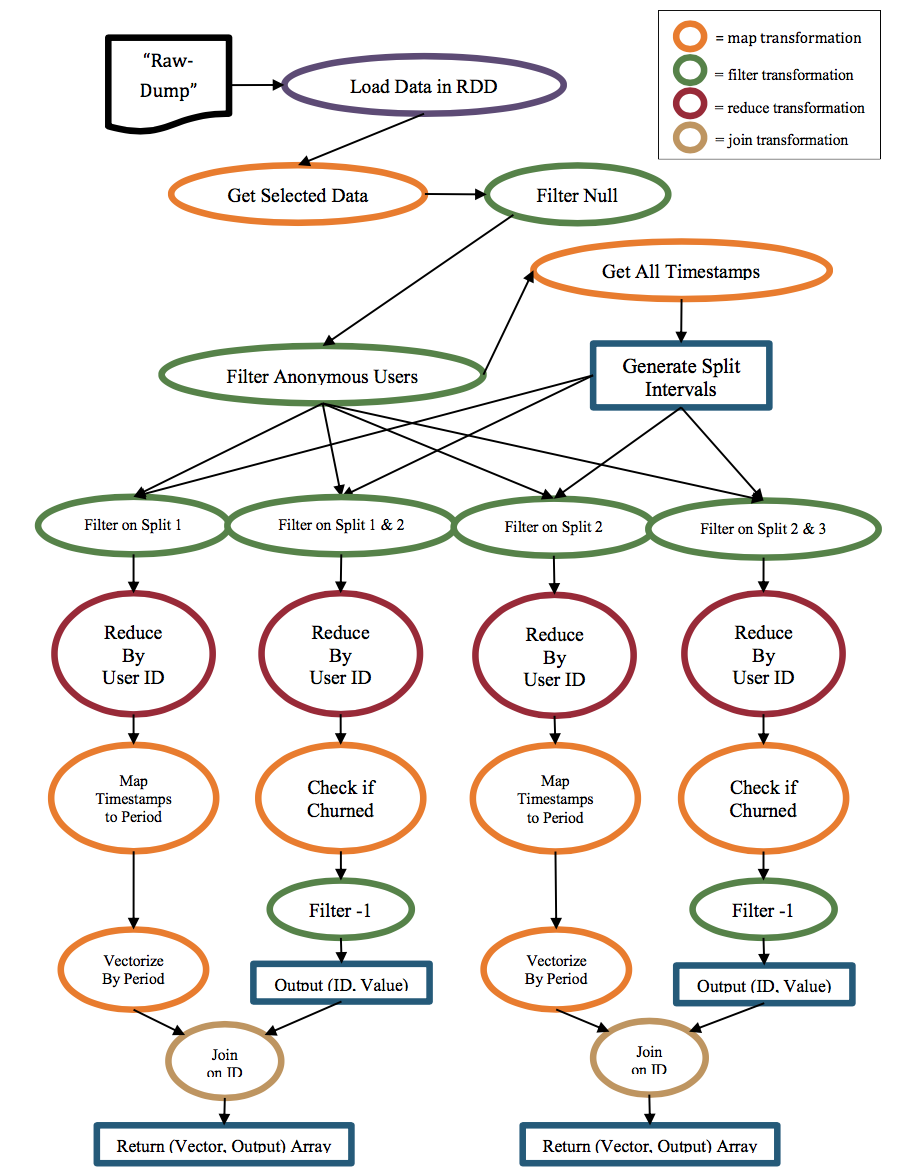}
\caption{Complete directed acyclic graph of the transformations performed in Spark in order to retrieve (Event Vector, Churn Output) mappings}
\end{figure}
\subsection{Implementing an Appropriate Deep Learning Architecture}
Implementing a deep network architecture can be impossible if attempted to be done without a way of expressing and computing mathematical expressions programmatically. Furthermore the programming language needs to be extremely fast in its computations as during the training phase of a neural network, a lot of different computations need to be performed (cost function calculations, gradient estimations etc.). Theano has long been recognised as an effective Python library in implementing deep neural networks, especially in research \cite{bastien}. It allows the definition, optimization and evaluation of mathematical expressions of arbitrary complexity. Once a mathematical function has be expressed and evaluation is initiated, Theano will compile the function into C code and automatically optimize the generated C code so that when it is evaluated, the computation is extremely fast. Furthermore Theano allows for GPU acceleration for its computations using the NVidia CUDA API. The advantage that ultimately led to the adoption of Theano in this project, was that the library can automatically perform differentiation on a function. This simplified the backpropagation algorithm greatly.\par
Theano uses special objects in order to effectively express any function. The most basic object is the Tensor object which essentially is a representation of the type of an expected input. For example let’s assume that the function $f(x)=x+y$ needed to be expressed in Theano. Variables $x$ and $y$ are defined as Tensor objects of type scalar. In other words $x$ and $y$ are expected to have integers as inputs. Subsequently variable $z$ can be defined as the operation to be performed on $x$ and $y$. Finally the complete function is expressed as having $x$ and $y$ as inputs and $z$ as the expected operation to be performed. Thus any function can be easily expressed by a combination of input Tensors and a single expression of the expected operation.\par
In order to better understand how Theano can be used to implement deep neural networks, several deep learning tutorials from the University of Montreal were implemented as practice (found here \cite{lisa}). These included logistic regression using a single artificial neuron up to the implementation of a Multilayer Perceptron (deep feed-forward architecture). The tutorials covered how Theano could be used to implement layers in a deep feed-forward network and how these layer objects could be easily stacked and trained using stochastic gradient descent on the popular MNIST dataset \cite{lecun:mnist}. Having realised the basic concepts of how various deep neural network mechanics could be implemented, it was decided that the deep learning architecture to be implemented would be based on the tutorial examples.\par
Since the only paper describing the application of deep learning in churn prediction, had proposed a deep feed-forward architecture, it was decided that such an architecture should be adopted. As mentioned in the paper, hyperbolic tangent activation neurons were used in their architecture. Having that said, background research suggested that the recently developed rectified linear activation neuron could allow for better backpropagation gradient estimations. Furthermore background research suggested the use of Dropout could allow for better generalization in deep architectures. Thus it was decided that both these mechanics would have to be implemented so that the final architecture could benefit from these techniques.\par
\begin{figure}[!t]
\centering
\includegraphics[width=3.5in]{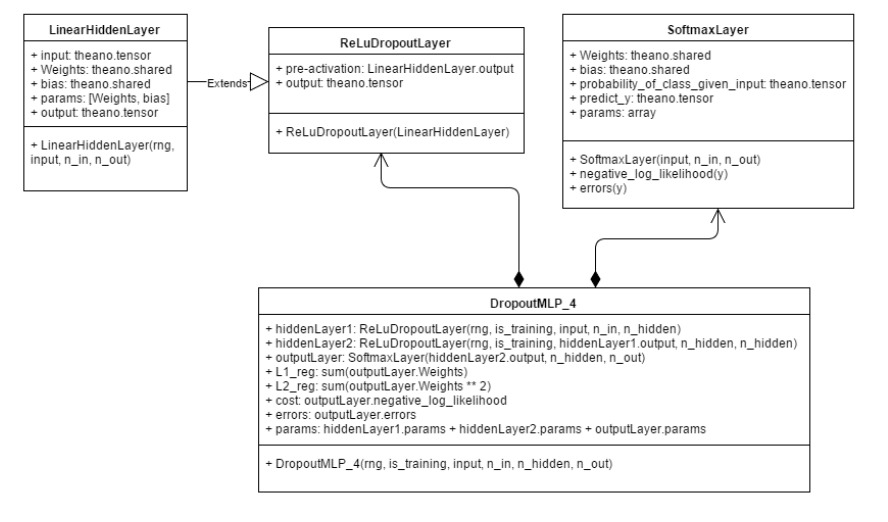}
\caption{UML Class diagram of the proposed architecture with 4 layers}
\end{figure}
After following the Theano tutorials, it was decided that four different classes needed to be created. The classes are depicted in the UML class diagram in Figure 17. The most basic class is the \emph{LinearHiddenLayer}, its main function is to generate the basic structure of a layer in an architecture using the \emph{“n\_in”} and \emph{“n\_out”} (neurons in, neurons out) arguments to generate an array of shape $N_{in} \times N_{out}$ and then assign the array values to a \emph{“Theano.shared”} variable. Shared variables allow information to be copied onto the GPU and provide access to their contents to all Theano functions, so that information is not constantly copied on the GPU in order to perform computations (severe decrease in performance). The array is randomly instantiated based on a random uniform distribution and essentially represents the weight values of a layer. The bias vector is instantiated in a similar way but instead, it is instantiated with zeros. Lastly the output parameter is expressed as the dot product between the input argument and the weights plus the bias.\par
The \emph{LinearHiddenLayer} class is extended into the ReLuDropoutLayer class whose basic function is to apply the rectified linear activation on the output parameter of the \emph{LinearHiddenLayer} subclass. The output of the \emph{ReLuDropoutLayer} class is expressed as a dropout function applied on the now activated layer (dropout function is discussed later on). \par
In addition to the dropout hidden layers, the output layer of the architecture required a $softmax$ layer as its output in order to be able to act as a classifier. A \emph{SoftmaxLayer} class was implemented to serve this purpose. Its weights and bias parameters are instantiated with zeros and similar to the \emph{LinearHiddenLayer} class are assigned to shared variables. The parameter \emph{“probability\_of\_class\_given\_input”} parameter is expressed through a \emph{“Theano.softmax”} function which takes the dot product of its weights and its input and during computation will return the probability of an input belonging to a certain class. In order for the architecture to be able to make predictions, the \emph{“predict\_y”} parameter is instantiated as an expression through the use of the \emph{“Theano.argmax”} function which during computation will return the index of the neuron which has the highest probability. Finally the \emph{SoftmaxLayer} class has two methods. The \emph{negative\_log\_likelihood} returns an expression of the architectures cost function based on the classes’ parameters and a given label vector $y$. The errors method returns an expression to compute the zero-one-loss of the layers prediction against a given label vector $y$.
This is all brought together under the \emph{DropoutMLP} class. The class example in Figure 17 demonstrates a \emph{DropoutMLP} class of 4 layers. The main function of the \emph{DropoutMLP} is to stack \emph{ReLuDropoutLayer} classes and at the end apply a \emph{SoftmaxLayer}. The first hidden layer takes as input the input vectors from the data representation architecture. Subsequently the second hidden layer takes the first hidden layers’ output as input and finally the output layer takes as input the second hidden layers’ output. It has to be noticed that the size of the hidden layers is intuitively assigned as a parameter by controlling the dimensions of each layers weight matrices. Since dropout is used in this architecture, L1 and L2 regularizations are applied only in the output layer and the values are respectfully computed as separate parameters. Finally the \emph{DropoutMLP} class takes the output layers negative log likelihood method and assigns the expression as the cost parameter. Similarly the errors parameter takes the output layers error method expression.\par
The \emph{DropoutMLP} object can now be effectively be trained by creating Theano functions that can train, validate and test the objects architecture using its public parameters in a backpropagation algorithm. The implemented backpropagation algorithm was essentially an altered stochastic gradient descent algorithm from the tutorial with an added momentum technique. Having that said, the Theano tutorials did not show how the dropout technique could be implemented or how the activation function for rectified linear neurons could be implemented. Furthermore the tutorials only demonstrated stochastic gradient descent without a momentum parameter. Thus these techniques needed to be implemented based on the research paper descriptions of each technique.\par
The rectified linear unit activation was implemented by expressing a function in Theano that could be applied on a layers pre-activation matrix (the dot product of the weight matrix and the input values) such that only the maximum of a Theano Tensor object would be returned. To implement dropout, a function was created that would allow only a proportion of the activations of a layer to be passed on to the next layer. The function can do this by generating an array of a randomized binomial distribution of 1 trial, of the same size of a layers activation matrix. The result of the binomial distribution is controlled by a parameter $p$ which corresponds to the probability of a neuron not dropping out (indicated by a 1 in the array). Thus the end result would be an array of ones and zeros of the exact same size as the activation matrix. By multiplying the array with the activation matrix, effectively only the results where a one is present will be returned. This function could be applied in any layers activation matrix and could demonstrate dropout. Having that said, dropout needs to be only applied during the training phase of a deep architecture. Simply adding this function to a layers class would not work as dropout will be constantly performed. Therefore a modification needed to be made in a dropout layers class such that dropout is only performed during training. This was done by using the Theano \emph{“Tensor.switch”} method which returns a variable depending on a conditions validity. By having the condition being \emph{“is\_training”} (Boolean variable) and by having that condition altered during the run time of the backpropagation algorithm, dropout could be switched on and off based on what phase the deep learning architecture was performing.\par
In order to implement momentum the expression of the \emph{“updates”} parameter of the Theano train function needed to be altered. The \emph{“updates”} parameter of the Theano train function essentially supplies the train function with an expression based on the weights of the architecture, which describes how the weights will be updated. Thus every time the train function is run, the weights of the architecture will be updated based on the result of the train function (cost) and the update expression supplied. Momentum can be incorporated in the update expression by firstly creating a Theano shared variable which essentially keeps track of each weight update across every iteration. Then the update expression was altered by incorporating a momentum parameter and having that parameter multiplied with the previous weight updates. This was subsequently added to the normal weight update expression where the learning rate is multiplied with the derivatives of the cost function with respect to each weight. As mentioned previously gradient derivation can be done automatically in Theano using the \emph{“Tensor.grad”} method by supplying it with the cost function and the weights. Furthermore during the end of every iteration the momentum parameter would need to be increased and kept under one while the learning rate would need to be decayed, so that correct gradient descent can be performed. This was done by instantiating two more Theano shared variables with the initial momentum and learning rate values. Therefore at the end of every iteration these shared variables were updated with their respective updated values. Momentum was increased by 2\% after every iteration until it reached 0.99, while the learning rate was decayed by 1.5\% after every iteration. This is known as learning rate annealing in gradient descent algorithms with momentum, and it is said to guarantee convergence to a minimum \cite{orr}.\par
Therefore the final implemented architecture is a deep feed-forward neural network with rectified linear activations in its hidden layers. Furthermore the architecture has a $softmax$ implementation as its output layer composed of two neurons (one for each class). This layer is regularized using L1 and L2 regularization. The architecture also has a dropout technique implemented during its training which should allow for better generalization. Lastly the architecture is trained using an implementation of stochastic gradient descent with momentum and an early stopping technique to further fight against overfitting.\par
\section{Evaluation \& Analysis}
Before any evaluation could commence, data from different companies needed to be gathered. Three companies were randomly selected and 390 days’ worth of \emph{“raw-dump”} event files were collected from each. It was necessary to take this much data as the proposed architecture effectiveness needed to be tested across different months. Due to confidentiality reasons, these companies will not be named and will be denoted as $company_1$, $company_2$ and $company_3$. Each company’s data was passed through the proposed Spark data representation algorithm with the split length parameter assigned to 30 days. It was decided that the input vectors of each company should be based on 60 days instead of just 30 (less sparsity in input vectors), therefore two splits were taken to build the input vectors. The churn calculations that generate the churn output values were performed on subsequent 30 days (1 split) and 60 days (2 splits). This effectively allowed for different churn projection windows as companies had varying churn rates. A company with a low churn rate would not be able to produce enough churn positive samples in a small churn projection window. Generated datasets for projecting different company splits were collected from the Spark cluster and were evaluated.\par
In order to effectively evaluate how well the proposed architecture performed, it was compared against a simple feed forward architecture on all generated datasets. The simple feed-forward architecture employed hyperbolic tangent activation neurons in its hidden layers, without dropout implemented and without momentum implemented in its backpropagation algorithm. This was done in order to evaluate the effects, if any of these newly introduced techniques. Furthermore the number of layers for both architectures was increased during tests from 4 to 6, to evaluate the effect of adding more hidden layers. Both architectures had their learning rates incremented from 0.0001 to 0.01 across all datasets in order to adjust for different error surfaces. Different datasets would have different error surfaces and because learning rate plays such an important role in determining an optimal minimum, it would not be wise to keep it constant for all datasets. The momentum parameter in the proposed architecture was kept constant across all tests with an initial value of 0.5 and was decayed at a constant rate. Finally the proposed architecture’s performance was compared against Framed’s current Random Forest algorithm across the same split intervals of the generated datasets.\par
\subsection{Evaluating the Effectiveness of Introduced Techniques}
\subsubsection{Company 1}
Through the first company’s data, it was possible to generate two sets of datasets with different churn prediction window timeframes, as enough churn samples were generated in both cases. Due to the fact that two splits (60 days) were used to generate the input vectors, the first split that could be predicted was the third one. In total ten splits could be predicted when the churn projection window was set to 30 days and nine splits when it was set to 60 days.\par
\begin{figure}[!t]
\centering
\includegraphics[width=3.5in]{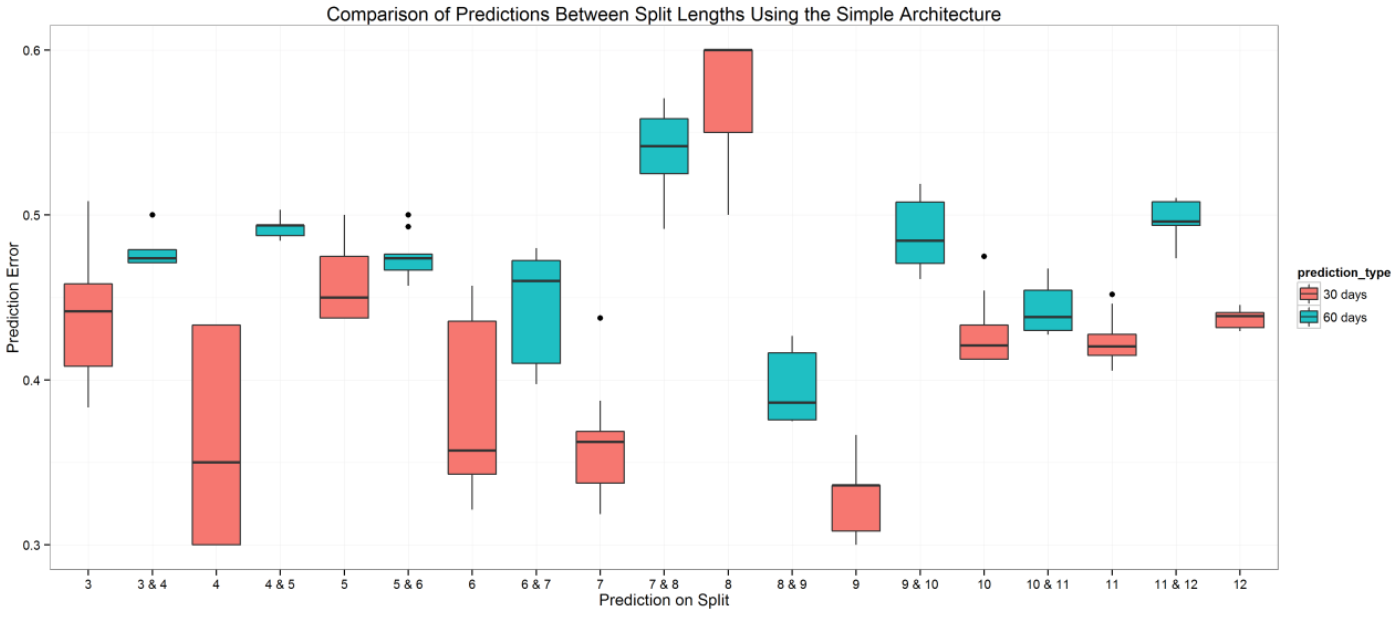}
\caption{Prediction Error of the Simple Feed-Forward Architecture across $company_1$ splits}
\end{figure}
The simple feed-forward architecture was tested across all generated datasets with varied learning rates and number of layers. The results can be seen in the box plot depicted in Figure 18. The length of each box can be expressed as the variance between prediction errors in terms of learning rate and layer numbers. It can be seen that when comparing the results between the two churn projection windows, the architecture performs much better in 30 day churn prediction timeframes. As more information exists in the input vectors of the larger prediction window, the results seem counterintuitive (more information should lead to less sparse input vectors and therefore better results). This might be an indication that too much information might “hide” the variances between the input vector positions and thus ultimately produce less separable data. Furthermore, when comparing the variance between the smaller churn prediction timeframe and the longer one, the results of the smaller timeframe have a much larger variance between the boxes across splits, while the longer one seems to be more stable. The lowest point of the box plot can be thought of as the architecture setup with the best result. Thus it can be said that the simple architecture was able to generate adequate prediction results on split 4, 6, 7 and 9 with its best results being 70\%, 68\%, 68\% and 69\% accuracy respectively $accuracy = (1 - error) \times 100$.\par
\begin{figure}[!t]
\centering
\includegraphics[width=3.5in]{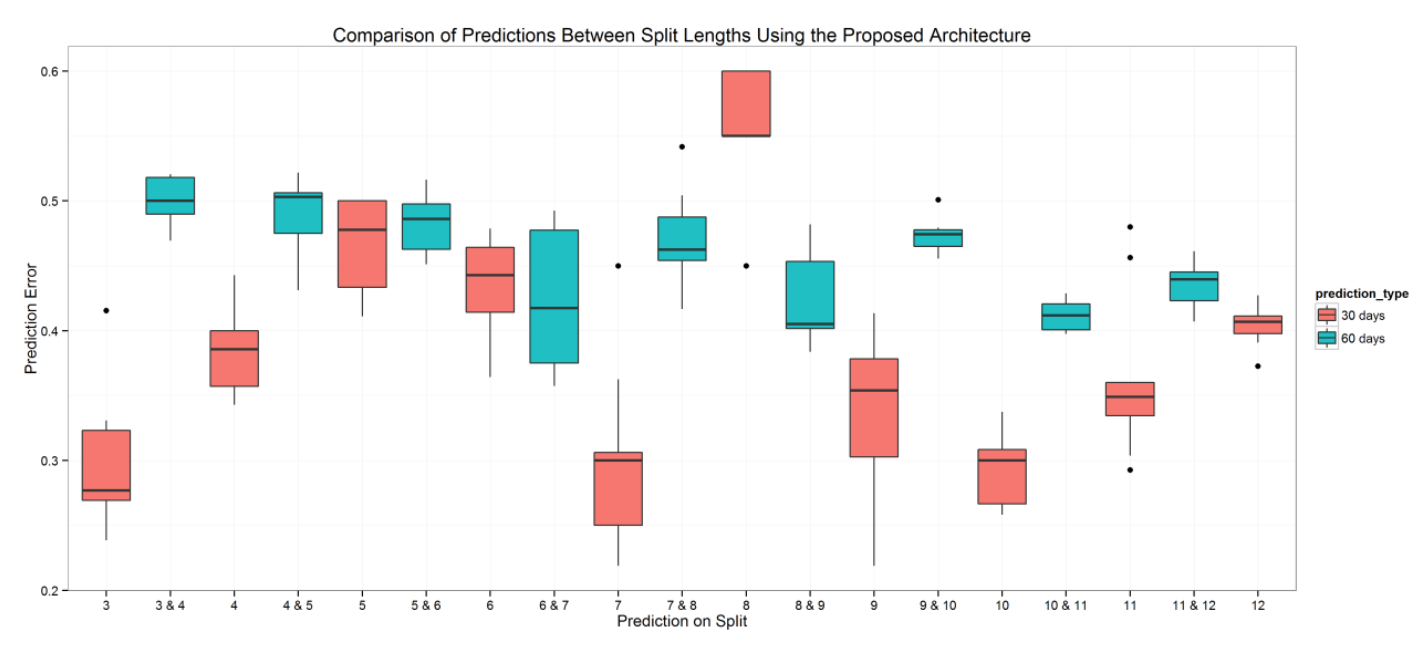}
\caption{Prediction Error of the Proposed Feed-Forward Architecture across $company_1$ splits}
\end{figure}
The proposed feed-forward architecture was subsequently tested on the same datasets. It was able to overall perform much better across all splits, with even some splits having prediction errors lower than 0.3 (Figure 19). Therefore it can be said that the proposed architectures activations could better decompose the variance between the feature vectors. The most interesting result is how the variance between the 60 day churn prediction window boxes has all but diminished. It could be said that this effect proves that the architecture demonstrates better generalization when compared to the simple architecture on the same splits. Even though the simple architecture performed better on split \emph{“8 \& 9”} it could be that it was overfitting the hypothesis function as other 60 day splits performed much poorly. Overall the introduced techniques in the proposed architecture seem to have a beneficial effect on the prediction results as the architecture demonstrates better generalization and much better accuracies across splits. Its best results where on splits 3, 7, 9, 10 with accuracies of 74\%, 78\%, 72\% and 75\% respectively.\par
\subsubsection{Company 2}
Similarly as with $company_1$, it was possible to generate enough churn samples for the two different sized churn prediction windows.
\begin{figure}[!t]
\centering
\includegraphics[width=3.5in]{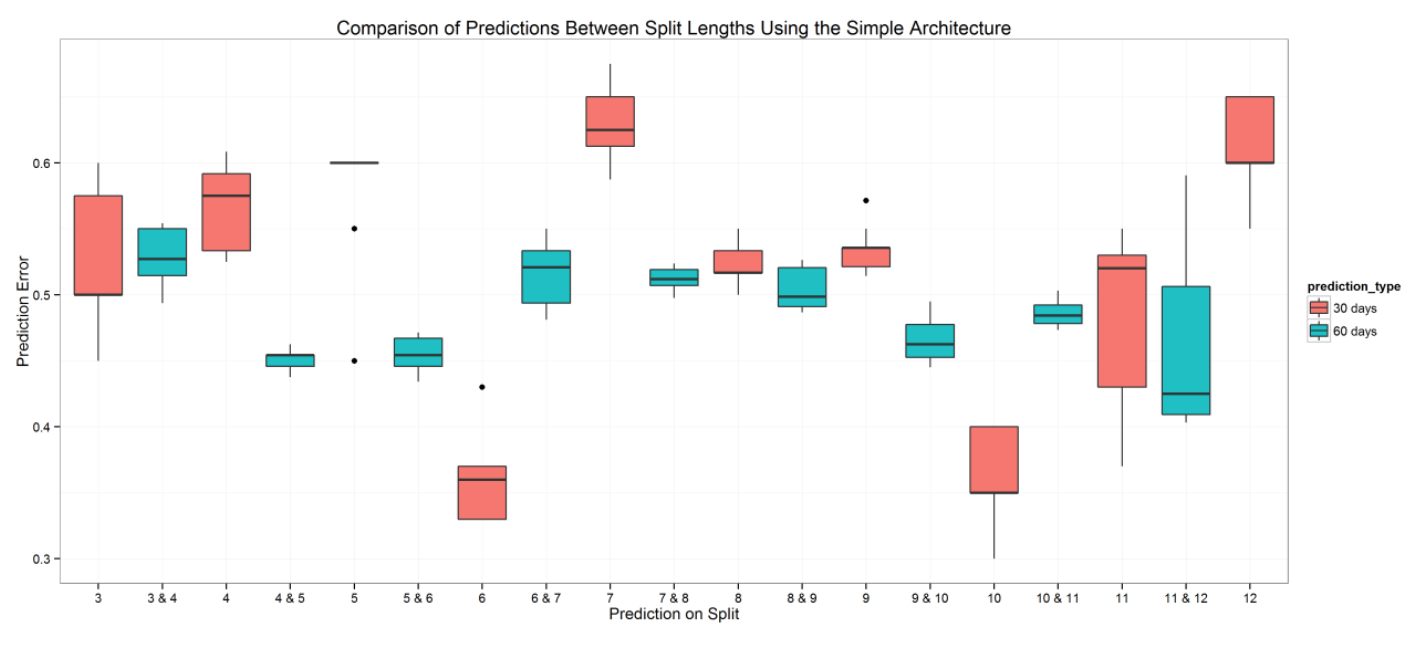}
\caption{Prediction Error of the Simple Feed-Forward Architecture across $company_2$ splits}
\end{figure}
Overall the results seem to be significantly worse when compared to the results obtained by the simple architecture across $company_1$ splits (Figure 20). The only reasonable prediction results were obtained on spits 6 and 10 and even then, the results weren’t particularly great with the best accuracies being 68\% and 65\% respectively. Variances between the boxes of the smaller churn prediction window are extremely high, showing that the architecture has difficulty generalizing across different months. Contrary to $company_1$ results it cannot be said that the difference in churn prediction windows had an immediate effect on the results, as most of the smaller prediction window results have similar results to the larger ones.\par
\begin{figure}[!t]
\centering
\includegraphics[width=3.5in]{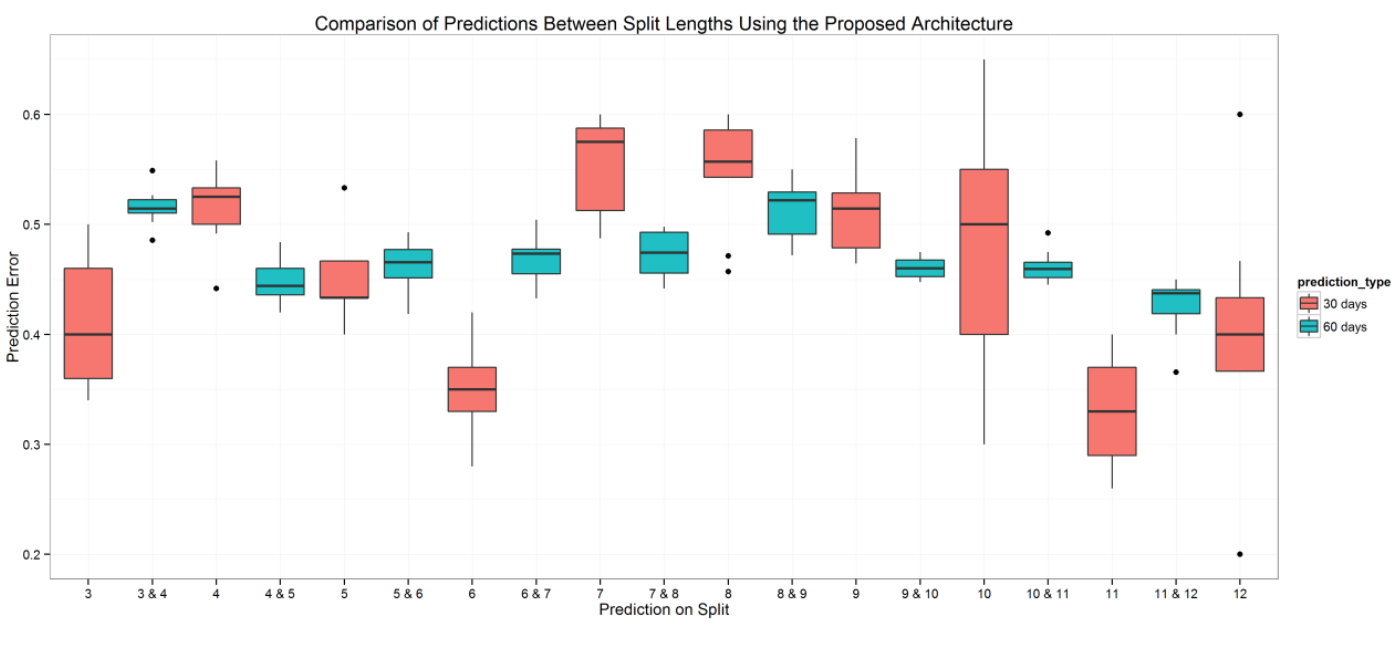}
\caption{Prediction Error of the Proposed Feed-Forward Architecture across $company_2$ splits}
\end{figure}
Testing the proposed architecture on $company_2$ datasets produced similar results as the simple architecture, with no significant increases in prediction accuracies across splits. Having that said, Figure 21 demonstrates how effective the techniques employed in the architecture are at creating more generalized hypothesis functions. Variances between the smaller projection window results have significantly decreased and variances between the larger projection window results have almost diminished. Although this does not explain why predictions on all datasets for both the simple architecture and the proposed one are so poor. This may be due to the proposed data representation, failing to capture any significant differences in patterns between churners and non-churners (non-separable data). Thus further investigation would have to be carried out to prove that this is the cause and to subsequently understand what might be the reason (discussed in subsection D).
\subsubsection{Company 3}
Data from $company_3$ could not produce an adequate amount of churn samples for a single split (30 day) churn prediction window. This was due to the low churn rate of the selected company and therefore only datasets based on two split churn prediction windows were generated.\par
\begin{figure}[!t]
\centering
\includegraphics[width=3.5in]{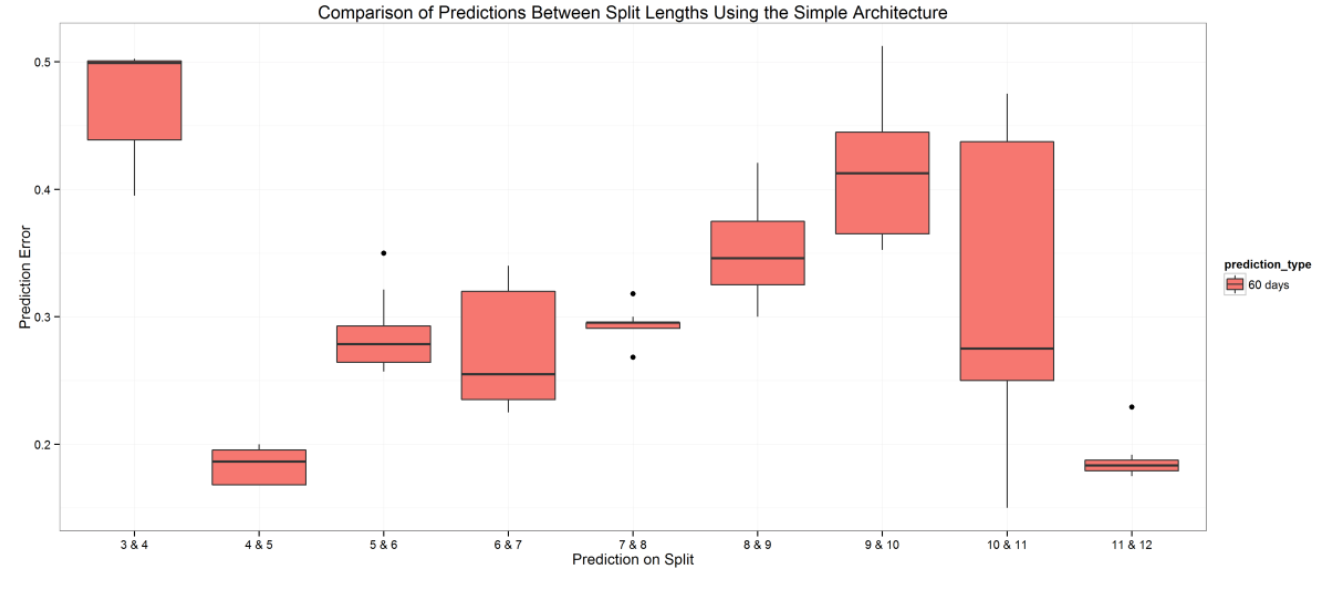}
\caption{Prediction Error of the Simple Feed-Forward Architecture across $company_3$ splits}
\end{figure}
As can be seen in Figure 22, the results of the simple feed-forward architecture again show high variances between splits, proving again that the architecture is not stable across different months. Having that said, the results overall are extremely better than the prediction results obtained in data from $company_1$ and $company_2$. More than half of the splits (at a specific architecture layer setup), produced accuracies higher than 70\% and in splits \emph{“4 \& 5”} and \emph{“11 \& 12”}, accuracies higher than 80\% were attained.\par
\begin{figure}[!t]
\centering
\includegraphics[width=3.5in]{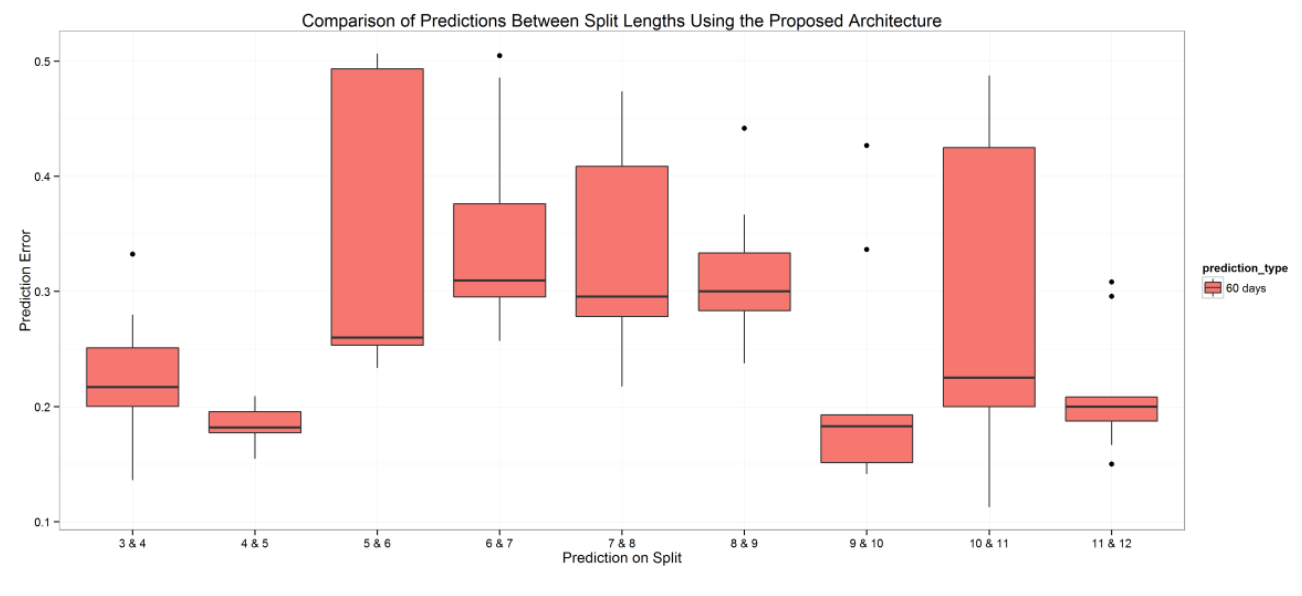}
\caption{Prediction Error of the Proposed Feed-Forward Architecture across $company_3$ splits}
\end{figure}
The results become even better when the proposed architecture is tested on the generated datasets of $company_3$ (Figure 23). At specific number of layer setups the architecture produces accuracies higher than 70\% across all of the splits. Furthermore four splits produced accuracies greater or equal to 80\% which is a significantly better result when compared to the results obtained from the simple feed-forward architecture. Most importantly the variances between split results have decreased in comparison to the variances in the simple architecture. Although it has to be noted that the individual prediction results of splits \emph{“5 \& 6”} and \emph{“10 \& 11”} show extreme fluctuations (length of the boxes). Further investigation (see subsection B) showed that these large variances were caused by the learning rate parameter assignment rather than the number of layers of a specific architectures’ setup. This of course demonstrates how important the learning rate parameter is to avoiding falling into suboptimal local minima, even when momentum is implemented in the gradient descent algorithm.\par
\subsection{Evaluating the Effect of Adding More Layers}
The results shown in the previous section were based on cumulative prediction performances of architecture setups with varying number of layers and varying learning rate assignments. In order to identify which of the two varying criterions played a more significant role on the prediction performances and to evaluate if increasing the number of layers produced an effect on prediction performance, the following plots were generated.\par
\subsubsection{Company 1}
\begin{figure}[!t]
\centering
\includegraphics[width=3.5in]{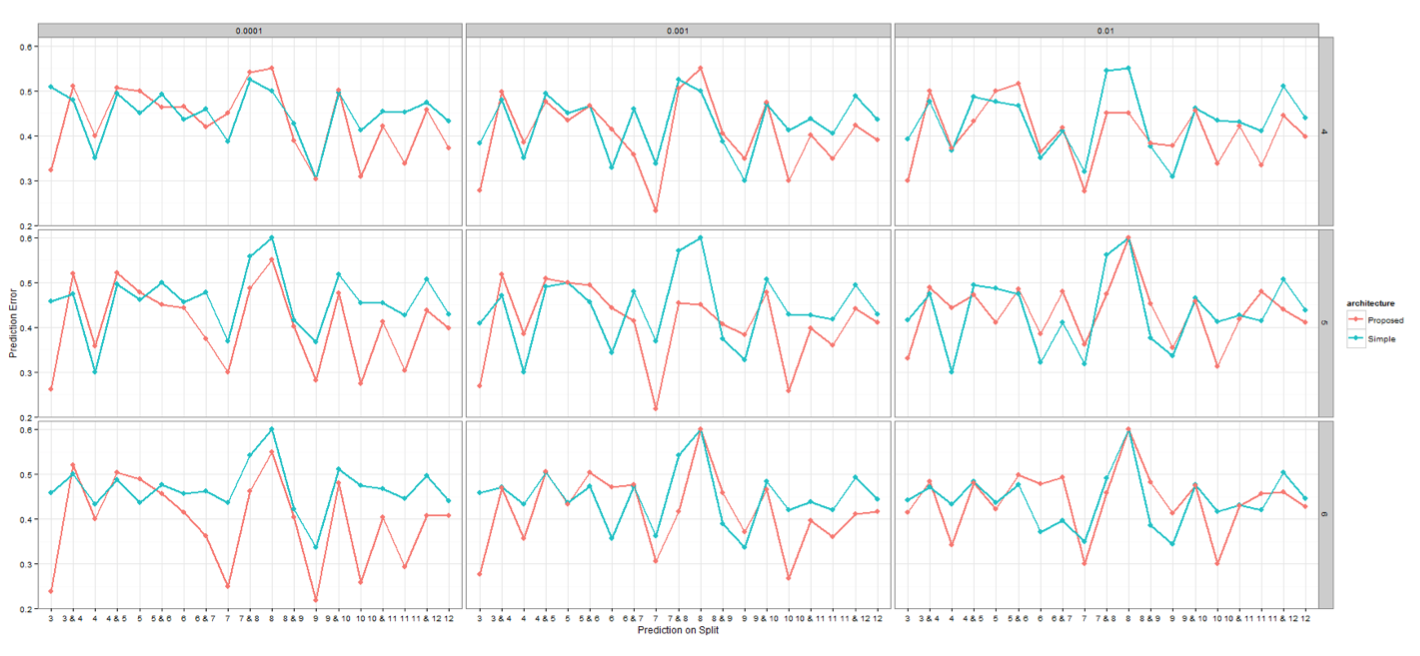}
\caption{Comparison of Learning Rate vs Number of Layers: Prediction Errors across all splits of $company_1$}
\end{figure}
The plot in Figure 24 depicts the effects the learning rate and the number of layers have in terms of prediction accuracies across all splits of $company_1$ between each of the architectures. The proposed architectures prediction errors are plotted in a red line and the errors of the simple feed-forward architecture are plotted in blue. The plot is split in sections with the learning rate varied from 0.0001 to 0.01 across the x-axis and the number of layers varied from 4 to 6 down the y-axis. Immediately it can be seen that adding more layers has little to no effect on the simple feed-forward architectures’ prediction results. On the other hand an increase in layers in the proposed architecture seems to allow the architecture to take advantage of the extra layers in order to generate more abstract features. This is clearly seen in the plot depicting both architectures with 6 layers and trained on the very small learning rate of 0.0001.\par
Viewing the figure from left to right will demonstrate the effect the learning rate has on each architecture setup. While comparing the two architectures this way, it can be seen that altering the learning rate has little to no effect on the employed stochastic gradient descent algorithm in the simple architecture. The proposed architectures’ added momentum parameter seems to be very dependent on what learning rate is chosen. From the plot it can be concluded that the employed backpropagation algorithm with momentum, overall seems to perform much better when small learning rates are used (0.0001 and 0.001). Lastly if an optimal setup would have to be suggested for predicting $company_1$ data, it would have to be a six layer proposed architecture with 0.0001 learning rate and 0.5 momentum as its hyper parameters. This is because it is the setup that produced the best results across most of the months. \par
\subsubsection{Company 2}
\begin{figure}[!t]
\centering
\includegraphics[width=3.5in]{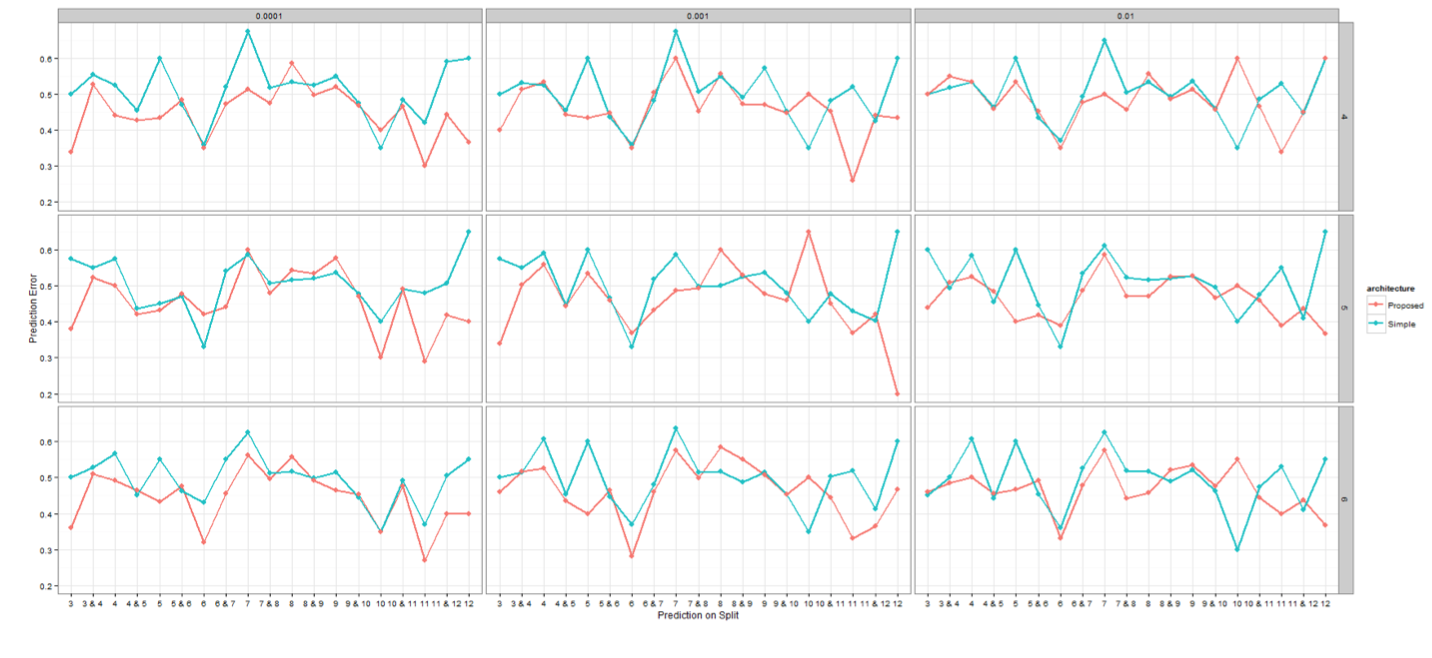}
\caption{Comparison of Learning Rate vs Number of Layers: Prediction Errors across all splits of $company_2$}
\end{figure}
A similar plot was generated for $company_2$ in Figure 25. Compared to the plot of $company_1$, the effects of both varying the learning rate and varying the number of layers are negligible. The results are counter intuitive which further promotes the concept that the problem with $company_2$’s results lie within the proposed representation architectures’ ability of effectively capturing the differences between churn and non-churn user event patterns. Although if an optimal setup would have to be suggested, it would be a proposed architecture of 6 layers trained using 0.0001 learning rate and 0.5 momentum as its hyper parameters, as it produced the best results across most of the splits.\par
\subsubsection{Company 3}
\begin{figure}[!t]
\centering
\includegraphics[width=3.5in]{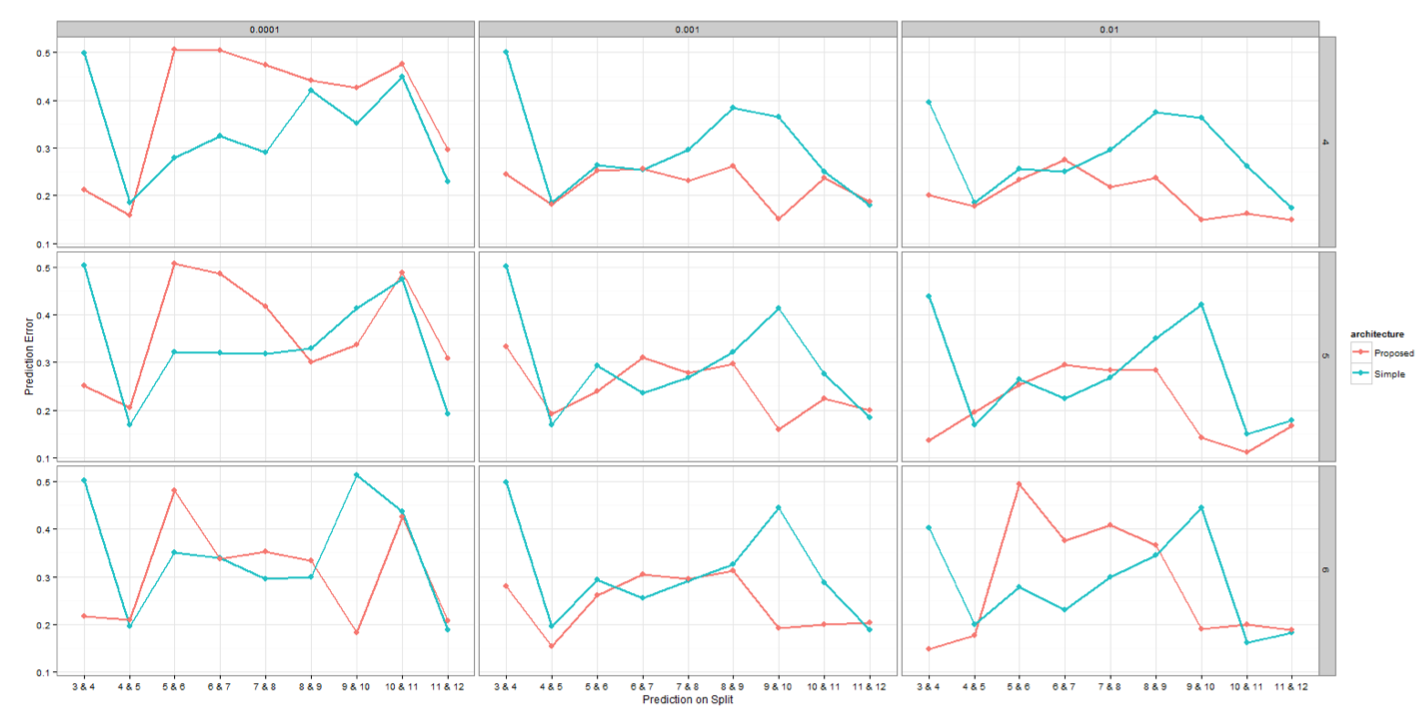}
\caption{Comparison of Learning Rate vs Number of Layers: Prediction Errors across all splits of $company_3$}
\end{figure}
As mentioned previously the boxplots generated for the proposed architecture in the previous subsection for $company_3$, showed high variance between prediction results on splits \emph{“5 \& 6”} and \emph{“10 \& 11”}. This effect was attributed to learning rate selection and this can clearly be seen in the plot in Figure 26. Viewing the plot from left to right, in every proposed architecture setup (number of layers) there seems to be a direct dependency on the selected learning rate and the prediction results. The very low learning rate of 0.0001 produced consistently poor results across most of the splits. Furthermore the highest learning rate seems to miss the minimum in most splits at the highest number of layer configuration. This could mean that the error surfaces of most of $company_3$ splits are riddled with a lot of local minima. \par
Due to the early stopping mechanism employed, the lowest learning rate could have never got the chance to reach a minimum. The largest learning rate could have consistently missed the minimum due to its large step size on the higher dimensional error space caused by the larger amount of layers. Furthermore considering the increase in layer numbers, $company_3$’s data seems to not benefit substantially from a larger amount layers, as was noticed with $company_1$. This might be due to the fact that the minimum number of layers was sufficient to defragment the complexities within the input vectors. Overall if an optimal architecture configuration would have to be suggested for predicting $company_3$ splits, it would have to be a proposed architecture of 4 layers trained using 0.001 learning rate and 0.5 momentum as its hyper parameters.\par
\subsection{Evaluating the Overall Performance against Frameds’ Model}
Even though the proposed representation architecture and the proposed deep learning architecture have shown promise in their applications, it is essential to see how well they perform against Frameds’ current Random Forest model. As mentioned in the objectives section, beating Framed’s model is not a requirement due to the fact that the general aim of the project is to demonstrate how the feature engineering step could be skipped. Having that said, if the proposed pipeline performs significantly worse than the Random Forests model, even if the feature engineering step is skipped, Framed would not look into integrating it in their pipeline.\par
The metric that was used to keep the comparisons between the two models as fair as possible, was the zero-one-loss metric. This is the same metric that the prediction results of the proposed deep architecture were based on. Furthermore it also has to be mentioned that Framed’s model is an implementation of Balanced Random Forests which means that it does not require any balancing in its datasets in order to correctly classify between classes of an imbalanced dataset. Thus Frameds’ model would essentially have more samples to work with than the proposed deep architecture, as dataset balancing is performed using random under-sampling. Lastly Frameds’ model would be compared to the “optimal” deep learning architecture configurations for each company (discussed previously).\par
\begin{figure}[!t]
\centering
\includegraphics[width=3.5in]{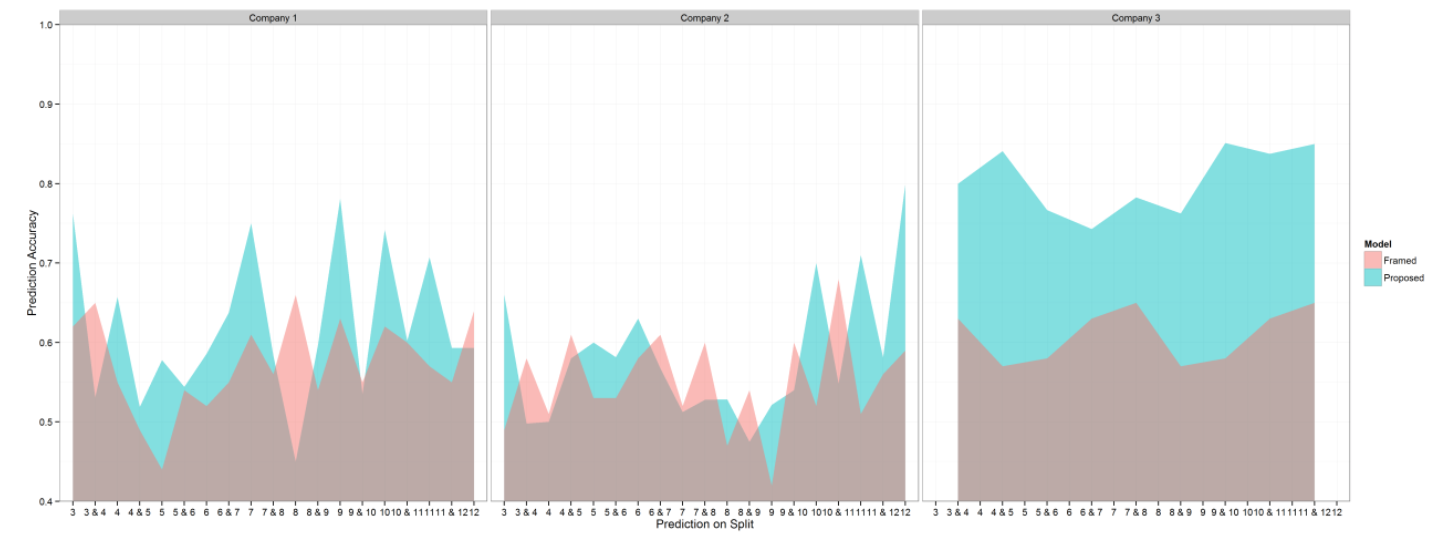}
\caption{Prediction accuracy comparison between employed model at Framed (red) and proposed model (blue) across all companies.}
\end{figure}
The plot shown in Figure 27, shows the prediction results of both Frameds’ model and the best performing proposed deep learning architecture configuration across all splits of each company. The Random Forests model prediction accuracies are depicted in red and the proposed deep architecture is depicted in blue. The prediction results are overlaid so that their differences can be clearly seen over each split.\par
In $company_1$’s prediction results it can be clearly seen that the proposed deep learning pipeline outperforms the Random Forests algorithm over almost every split (except three). Furthermore in some of the splits the proposed deep learning pipeline exceeds the Random Forests’ performance with an increase of almost 15\%, which is a remarkable increase in predictive performance. Framed’s Random Forests algorithm outperforms the proposed architecture substantially on split 8. The reason behind this was identified to be that the particular splits’ dataset only had 22 samples in its validation set. This was caused by an exceedingly low number of churners during the splits timeframe, thus when the dataset was randomly under sampled for balancing, most of the splits samples were removed. Therefore the proposed deep architecture was not able to correctly train its weights on such a small validation set.\par
Contrary to what was observed in $company_1$'s predictions, the prediction results for $company_2$ showed the proposed deep learning pipeline was not consistently beating the performance of the Random Forests model. Interestingly almost all the splits that the Random Forests model was able to beat the proposed deep architecture, were splits of 60 day churn prediction windows. This could be caused by the way the input vectors of the proposed data representation could “lose” variance between each vector position such that discriminatory patterns are lost.\par
The most promising results for the proposed deep learning pipeline, are the result comparisons on $company_3$ Throughout all the splits, the deep learning pipeline outperforms the Random Forests model by an exceptional average prediction accuracy difference of 20\%. Furthermore, considering the fact that the deep learning architecture was trained on datasets with no manual feature engineering, the results of $company_3$ overall show great potential of an adoption of a deep learning pipeline at Framed could significantly increase prediction performance and while excluding the feature engineering stage.\par
\subsection{Examining the reason behind Company 2’s poor performance}
The prediction results of $company_2$ demonstrated consistent poor performance across almost all of its splits when compared to $company_1$ and $company_3$ results. It was previously mentioned that the suspected reason behind this was that the proposed data representation architecture could not effectively capture the differences between user event vectors. In order to test this theory, training set samples were taken from each company and passed through a nonlinear dimensionality reduction technique known as t- distributed stochastic neighbour embedding (t-SNE) \cite{maaten}, in order to determine if the data of each company was able to separated. The technique has been extensively used to visualize the structure of high dimensional datasets.\par
\begin{figure}[!t]
\centering
\includegraphics[width=3.5in]{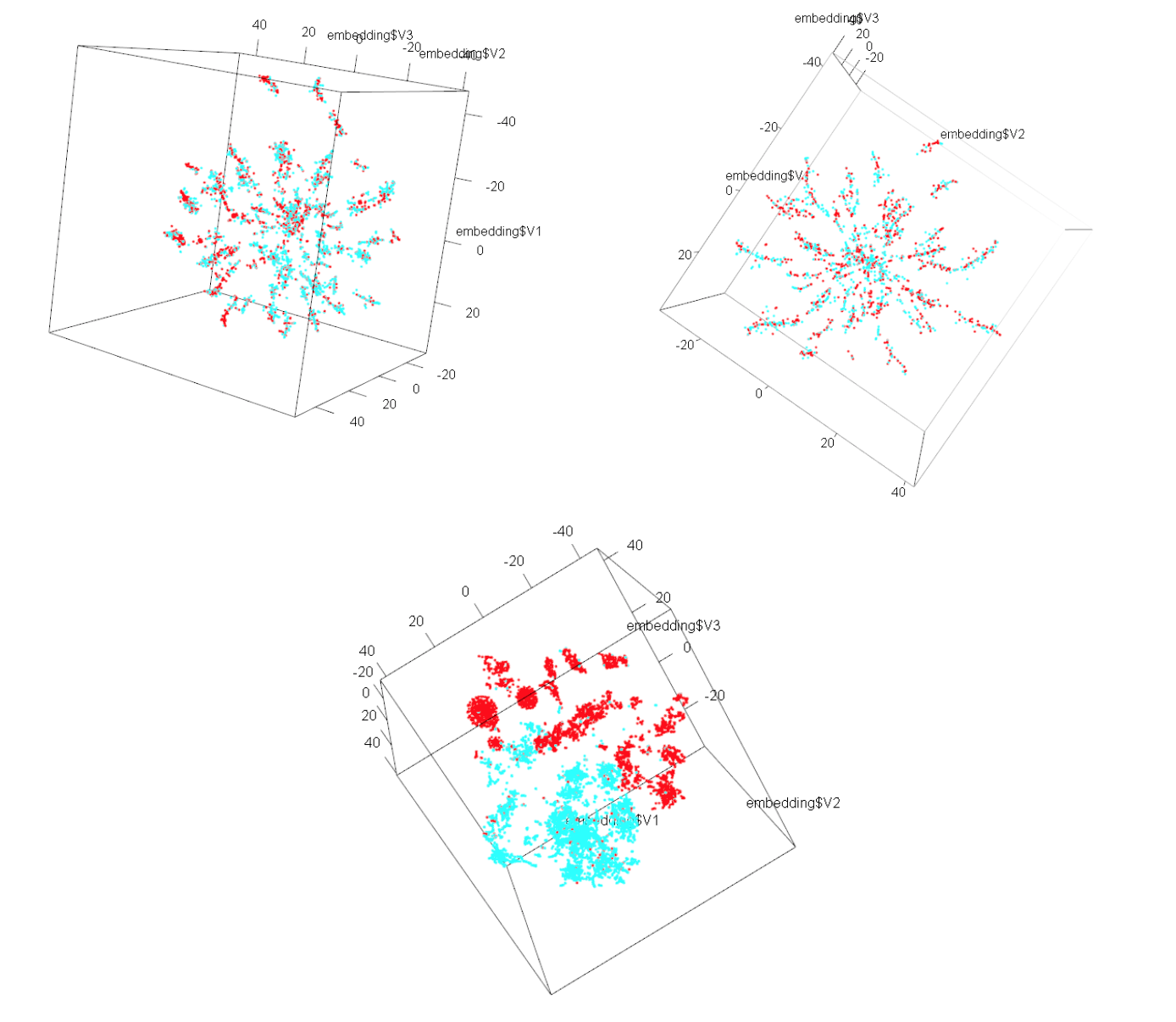}
\caption{t-SNE Visualizations of company train datasets. Top Left: $company_1$ Top Right: $company_2$, Bottom: $company_3$}
\end{figure}
Examples of the dimensionality reduction results on different datasets, can be seen in Figure 28. The results are based on training sets from each company that had the largest amount of samples. Immediately, it can be noticed that $company_3$ training data is easily separable which in turn would explain the remarkable prediction results. When looking at the data separation of $company_1$, a slight separation can be seen on the left hand side of the plot. Even though the separation is not clear, through the training of the proposed deep learning architecture more abstract features were created such that the architecture was able to separate the data much better. This would explain why the performance of $company_1$ data seemed to perform better with the maximum amount of layers in the proposed architecture. Unfortunately the results of the dimensionality reduction on $company_2$ proved that the data has no apparent separation between its classes, which confirms the initial theory that the data representation was not able to capture pattern differences between the classes. This also explains why the best performance across the splits was found to be through the use of the 6 layer proposed architecture, as more layers would increase the performance if adequate abstract features are created during training.\par
Framed was approached with the evaluation results and they subsequently suggested that the only difference between $company_2$ and the other two companies was that it was not a subscription based company. In other words its registered users would not be paying on a monthly basis for a particular service. This was interesting information, as a non- subscription company would generally not have continuous customer event triggers, as customers would not feel the need to use the service frequently as they are not paying for it. Furthermore due to the logic behind the proposed data representation in measuring even counts differences over time, this would explain why it was not effective at capturing differences between churners and non-churners.\par
\section{Discussion}
The project investigated the hypothesis that an abstract, company independent data representation could be developed and used to train deep learning architecture in the problem of churn prediction. Through a deep learning architectures’ inherent ability of creating more abstract features though it’s hidden layers it was hoped that the data representation could provide adequate prediction results. \par
Looking back at the results of the evaluation and analysis section, the results show the great potential of the overall proposed deep learning pipeline to pose as a solution to the stated hypothesis. The proposed data representation architecture is able to capture customer event patterns through the use of very abstract information that should be available in any company data that logs user events. Furthermore the data representation architecture applies company independent logic to ascertain whether a user has churned based on a user’s inactivity. This was proved by visualizing the generated representations through the nonlinear dimensionality reduction technique t-SNE on $company_3$ and $company_1$.\par
Results revealed that the developed representation does not work well on non- subscription based companies. The developed data representation was designed with the assumption that once a user becomes inactive for 30 consecutive days, he becomes a churner. In subscription based companies this works particularly well as users of a service will feel less inclined to pay for a service if they are not using it, which in turn causes them to churn. This effect is demonstrated through the t-SNE visualization of $company_2$ where the data representation could not capture the differences between churners and non- churners. Thus the key mistake was to assume that all of Frameds' customers were subscription based.\par
Even with this set back, the proposed deep feed-forward architecture performed exceptionally well even on data representations where the data wasn’t particularly separable. This has been demonstrated by the prediction results of $company_1$ whose t-SNE visualization showed a very mild separation between classes. The proposed architecture was able to generate more abstract data features across its hidden layers which in turn allowed the architecture to better distinguish the differences between churner and non- churner input vectors. Furthermore this effect was shown to get better as the layer numbers were increased.\par
In addition to its ability of generating abstract features, the proposed deep architecture employed techniques that allowed it to generalize its hypothesis functions better across different splits. This was established by comparing it against a simple feed-forward architecture with hyperbolic tangent activations and no generalization techniques apart from L1 and L2 regularizations. Through the use of dropout the proposed architecture was able to inherit ensemble classifier traits that allowed it to produce less varied results across its months. Furthermore by employing rectified linear activations and momentum in its backpropagation algorithm, the proposed architecture demonstrated much better prediction accuracies than the simple feed-forward architecture.\par
Lastly the complete proposed pipeline (data representation architecture and deep feed- forward architecture), overall produced better prediction results than the currently employed machine learning algorithm at Framed. This was shown by comparing both models by the same metric and on the same split timeframes. Having that said, the employed dataset balancing technique proved to produce problems on splits with low churn samples. This in effect caused the proposed deep architecture to significantly underperform against Framed’s Random Forest algorithm. Therefore it can be said that the proposed pipeline is vulnerable on months with low churn rates. \par
\section{Conclusion}
Taking everything into account it can be said that the general aim of the project has been achieved. The project investigated and developed a representation architecture that could be applied to an arbitrary company that is able to log user events. Furthermore through prediction results it has been proved that it is effective at reducing the dimensionality of incoming data while being able capture a pattern representations of the underlying data features. Even if its effectiveness depends on what business model a company employs (subscription based etc.) the overall prediction results showed that through a deep learning architecture, the data representation does not underperform when compared to the currently employed feature engineering methods at Framed. In retrospect further companies should have been tested in order to better understand its effectiveness.\par
Furthermore through the development of the representation architecture, a cluster computing technology was implemented so that the generation of the proposed data representations could be realised. This took up a considerable amount of time away from exploring the data representation architecture further and other deep learning architectures. Having that said, the development of these technologies made learning and understanding of these technologies possible. Moreover the utilization of these technologies have allowed Framed to realize their potential and have expressed an interest in their adoption. As this was a research project for Framed, the development of these technologies can be thought of as an extension to the overall research performed.\par
The project also investigated and implemented an appropriate deep learning architecture that can demonstrate unsupervised feature learning and can be applied in the problem of churn prediction. In addition to the above objectives, the project incorporated modern deep learning concepts which greatly benefited the overall performance of the model.\par
Being newly introduced to deep learning, in depth research and comprehension of the underlying mechanics needed to be covered before any implementation could be performed. Initially further architecture types were planned to be investigated (Recurrent Neural Networks, Deep Belief Networks etc.), but due to the time constraints and the background understanding that needed to be covered, this proved to be unrealistic. The addition of an unsupervised generative pre-process architecture, like Deep Belief Networks, behind the proposed architecture, might have greatly improved the prediction results. Furthermore Recurrent Neural Networks are excellent in generating predictions based on temporal data, which is exactly what churn prediction is. Thus both of these architectures could be investigated further in future work. \par
\section{Acknowledgments}
The authors would like to acknowledge Elliot Block, Andrew Berls, Dan Evans, and the rest of the Framed Data team for their engineering and data infrastructure assistance and technical guidance. Simon Tomlinson from the Data Science Institute at the University of Lancaster also provided additional guidance and advice.
\ifCLASSOPTIONcaptionsoff
  \newpage
\fi

\end{document}